\documentclass{article}
\PassOptionsToPackage{compress,round}{natbib}
\usepackage[main,final]{neurips_2026}
\usepackage[utf8]{inputenc}
\usepackage[T1]{fontenc}
\usepackage[x11names,table,xcdraw]{xcolor}
\usepackage{booktabs,multirow,makecell}
\usepackage{amsmath,amssymb,amsfonts,mathtools,amsthm}
\usepackage{nicefrac,microtype}
\usepackage{graphicx}
\usepackage{tikz}
\usetikzlibrary{arrows.meta}
\usepackage[labelfont=bf]{caption}
\usepackage{subcaption,wrapfig,float,placeins}
\usepackage[normalem]{ulem}
\usepackage{algorithm,algorithmic,siunitx}
\usepackage{hyperref,url}

\definecolor{lightgrayrow}{gray}{0.95}
\newcommand{\hl}{\cellcolor{lightgrayrow}}
\definecolor{lightskybluerow}{RGB}{234,245,252}
\newcommand{\mainhl}{\cellcolor{lightskybluerow}}

\newcommand{\method}[1]{\textsc{#1}}
\newcommand{\AUM}            {\method{AUM}}
\newcommand{\DUAL}           {\method{DUAL}}
\newcommand{\DUALBeta}       {\method{DUAL+Beta}}
\newcommand{\ELFS}           {\method{ELFS}}
\newcommand{\ELFSDINO}       {\method{ELFS (DINO)}}
\newcommand{\ELFSSelfEnc}    {\method{ELFS (Self-Encoder)}}
\newcommand{\FullySup}       {\method{Fully Supervised}}
\newcommand{\Prototypicality}{\method{Prototypicality}}
\newcommand{\Random}         {\method{Random}}
\newcommand{\ScoreExtrap}    {\method{Score Extrapolation}}
\newcommand{\SemiAUM}        {\method{Semi-AUM}}
\newcommand{\SemiAUMCutoff}  {\method{Semi-AUM+Cutoff}}
\newcommand{\SemiDUAL}       {\method{Semi-DUAL}}
\newcommand{\SemiDUALBeta}   {\method{Semi-DUAL+Beta}}
\newcommand{\SemiPrune}      {\method{SemiPrune}}
\newcommand{\ZCore}          {\method{ZCore}}

\title{Label-Efficient Dataset Pruning \\ via Semi-Supervised Pseudo-Labeling}

\author{
Yeseul Cho$^{1}$\thanks{Work done at KAIST.},
Baekrok Shin$^{2}$,
Changmin Kang$^{2}$,
Chulhee Yun$^{2}$
\\
$^{1}$Nota Inc.,
$^{2}$Kim Jaechul Graduate School of AI, KAIST
\\
\texttt{yeseul.cho@nota.ai}
\\
\texttt{\{br.shin,cmkang8128,chulhee.yun\}@kaist.ac.kr}
}

\begin{document}
\addtocontents{toc}{\protect\setcounter{tocdepth}{-1}}
\maketitle
\begin{abstract}

Dataset pruning reduces the storage and training costs of deep learning by selecting an informative subset from a large dataset. However, most existing pruning methods require fully labeled data, which limits their applicability in realistic settings where unlabeled data are abundant and annotation is costly.
Recent label-free pruning methods address this issue, but they rely on features from pretrained models to estimate example difficulty. This dependence can be unreliable when the target dataset differs substantially from the pretraining distribution. 
We propose \textbf{\SemiPrune}, a label-efficient dataset pruning framework that uses a small randomly labeled subset and semi-supervised learning to generate pseudo-labels for unlabeled data, enabling existing supervised pruning methods to operate on the resulting pseudo-labeled training pool. We then estimate example difficulty from pseudo-label-induced training dynamics and select a coreset. 
By learning directly from the target dataset, our method better captures the target distribution and provides more reliable signals for difficulty estimation and coreset selection. We validate our approach on domain-specific, image-corrupted, and long-tailed datasets, where it achieves state-of-the-art performance among label-free and label-efficient baselines, while also demonstrating competitive performance on standard benchmarks.
\footnote{Code is available at \url{https://github.com/cyseul/SemiPrune}.}
\end{abstract}

\section{Introduction}

Modern deep learning relies heavily on massive datasets, resulting in substantial storage and training costs \citep{achiam2023gpt, touvron2023llama}. To address this challenge, dataset pruning aims to remove redundant examples while retaining the most informative subset of the dataset, namely the \textit{coreset}. Existing pruning methods have demonstrated strong performance when models are trained on the retained coreset \citep{Coleman2020Selection, paul2021deep, zheng2023coveragecentric, maharana2024mathbbd, he2024large, zhang2024spanning, cho2025lightweight}. However, most of these methods require an initial training stage on fully labeled data. Since fully labeled datasets are limited in practice, and annotating all candidate examples solely for pruning can be prohibitively expensive, developing pruning methods under a limited annotation budget is highly desirable.

Several label-free methods~\citep{sorscher2022beyond, zheng2025elfs, griffin2026zero} reduce this annotation cost by estimating example importance from features extracted by pretrained models. 
However, their reliability depends heavily on how well the pretrained feature geometry aligns with the target data. When the target distribution differs from that of the pretraining data, such as in domain-specific, image-corrupted, or class-imbalanced datasets, feature-based difficulty estimates may fail to reflect the true structure of the data.

In contrast, semi-supervised learning (SSL) can effectively learn the distribution of a target dataset using only a small budget of labeled examples~\citep{sohn2020fixmatch, wei2021theoretical}. Leveraging this ability, we propose \SemiPrune{}, a framework for
label-efficient dataset pruning that first generates pseudo-labels
through SSL and then applies existing supervised pruning methods.
Because the SSL model is trained directly on the target dataset,
its pseudo-labels better reflect the target class structure and
provide more reliable supervision for difficulty estimation.
This approach requires only a small fraction of examples to be
initially annotated, reducing the annotation cost of coreset
selection relative to fully supervised pruning.

We evaluate our framework in settings where target-specific difficulty estimation is particularly important. These include domain-specific datasets such as Food-101 and SUN397, corrupted variants of CIFAR-100 and Tiny-ImageNet, and long-tailed datasets including Caltech-101 and a long-tailed variant of CIFAR-100. Across these challenging settings, our method achieves state-of-the-art performance among pruning baselines that use either no labels or only a small fraction of labeled data. We further evaluate our method on standard benchmarks, including CIFAR-10, CIFAR-100, and ImageNet-1K, showing that it remains competitive even when pretrained features are already well aligned with the target data.

\section{Related Works}

\subsection{Dataset Pruning}

\paragraph{Fully-labeled approaches.}
Supervised pruning methods typically assign importance scores to
examples and retain those with the highest scores~\citep{toneva2018an,
Coleman2020Selection, paul2021deep, zhang2024spanning}.
However, selecting only the top-$k$ examples can reduce coverage of the
data distribution, substantially degrading performance at high pruning
ratios. To address this issue, several recent approaches have proposed sampling methods that can be combined with the scoring metric~\citep{xia2023moderate, zheng2023coveragecentric, maharana2024mathbbd, cho2025lightweight}.
Some methods instead formulate pruning as an optimization problem and solve it greedily under certain constraints \citep{borsos2020coresets, killamsetty2021grad}.

\paragraph{Label-free approaches.}
\Prototypicality{} \citep{sorscher2022beyond} selects examples based on their distance to cluster centroids in the embedding space learned by SWaV~\citep{caron2020unsupervised}. 
More recently, \ELFS{}~\citep{zheng2025elfs} generates pseudo-labels through deep clustering on DINO embeddings~\citep{adaloglou2023exploring,
caron2021emerging, oquab2024dinov}, trains a scoring model on the
pseudo-labeled pool, and selects a coreset using AUM-based double-end
pruning.
Another zero-shot pruning method, \ZCore{} \citep{griffin2026zero}, constructs a zero-shot embedding space using a foundation model \citep{radford2021learning} and then iteratively samples lower-dimensional embedding subspaces to identify examples that provide broad coverage of features while reducing redundancy. 

\paragraph{Few-label approaches.}
Recently, label-efficient pruning methods that require only a small fraction of labeled samples have also been proposed. \ScoreExtrap{}~\citep{schmidt2026effective} trains a scorer on a small
labeled subset and extrapolates importance scores to the remaining
examples using $k$-nearest neighbors and graph neural networks.
Our framework instead uses the initial labels for SSL-based
pseudo-label generation, enabling supervised scoring on the full pool.

\subsection{Pseudo-labeling for Unlabeled Data}

Pseudo-labeling extracts supervisory signals from unlabeled data, but its function differs depending on the available supervision. In fully unsupervised representation learning, pseudo-labels are often obtained through deep clustering, where examples are grouped in a learned feature space and the resulting cluster indices are used as training targets~\citep{caron2018deep, asano2020selflabelling, caron2020unsupervised, van2020scan}. Recent deep clustering methods further show that utilizing rich pretrained representations from self-supervised ViTs and vision-language models such as CLIP~\citep{radford2021learning}, DINO~\citep{caron2021emerging}, and DINOv2~\citep{oquab2023dinov2} substantially improves clustering quality on standard benchmarks~\citep{adaloglou2023exploring, cai2023semantic}. However, because these assignments are produced without class annotations, they do not necessarily correspond to semantic classes. Their primary role is instead to impose structure on the representation space by grouping examples with similar learned features.

In semi-supervised learning, pseudo-labeling serves a more class-directed purpose, as a small labeled subset anchors the predictions on unlabeled data. \citet{lee2013pseudo} introduces pseudo-labeling as a simple method that treats confident predictions on unlabeled examples as training targets. Modern consistency-based methods build on this idea by generating pseudo-labels from weakly augmented images and using them to supervise strongly augmented counterparts. FixMatch filters these predictions with a fixed high-confidence threshold~\citep{sohn2020fixmatch}, while later methods refine this process through adaptive thresholding~\citep{zhang2021flexmatch, wang2023freematch}. Since these pseudo-labels are guided by target-class supervision, they are more likely to approximate semantic classes and can propagate label information from labeled to unlabeled samples~\citep{wei2021theoretical}.

This distinction matters for dataset pruning because pseudo-labels
provide the supervision used to estimate example difficulty.
SSL offers a way to adapt this supervision to the target classes
using a small labeled subset.

\subsection{Orthogonal Works}

Active learning also aims to reduce annotation cost, but iteratively acquires labels for informative samples through oracle queries during training~\citep{ash2019deep, killamsetty2021grad, hacohen2022active}. Consequently, its sample selection is closely tied to the current state of the model and proceeds progressively over the training process. In contrast, our method starts from a fixed, small randomly selected labeled subset, uses it to train a model under an SSL objective, and then constructs a coreset in a one-shot manner based on the pseudo-labels generated by that model.

Our work also differs from dynamic dataset pruning methods~\citep{qin2024infobatch, okanovic2024repeated, yuan2025instancedependent, wu2025partial}. Dynamic pruning methods update the retained subset during training
and maintain access to the full data pool~\citep{qin2024infobatch,
okanovic2024repeated, yuan2025instancedependent, wu2025partial}.
We instead select a fixed coreset offline and train without revisiting
discarded examples.
Appendix~\ref{appendix:dynamic_pruning} compares these settings using
InfoBatch and Partial Forward Blocking under matched optimizer updates.

We also distinguish our method from pruning approaches designed for SSL frameworks~\citep{killamsetty2021retrieve}. Unlike these methods, which use the retained coreset within SSL training, we use SSL only to generate pseudo-labels and then train the model on the selected coreset under standard supervised learning.

\section{Methodology}
\label{sec:methodology}
\subsection{Method Overview}
\begin{figure}[t]
\centering
\begingroup
\definecolor{fwblue}{RGB}{45,104,145}
\definecolor{fwteal}{RGB}{44,126,119}
\definecolor{fwink}{RGB}{43,57,69}
\definecolor{fwline}{RGB}{112,131,144}
\definecolor{fwsky}{RGB}{234,245,252}
\definecolor{fwmint}{RGB}{237,247,244}

\newcommand{\fwData}[4]{%
  \begin{scope}[shift={(#1,#2)},yshift=.06cm,xscale=.78,yscale=.917647,line width=.55pt]
  \path[draw=fwink!80,fill=#3] (-.29,.20) -- (-.29,-.23)
    .. controls (-.29,-.36) and (.29,-.36) .. (.29,-.23)
    -- (.29,.20) -- cycle;
  \draw[draw=#4] (-.29,.05) .. controls (-.29,-.08) and (.29,-.08) .. (.29,.05);
  \draw[draw=#4] (-.29,-.10) .. controls (-.29,-.23) and (.29,-.23) .. (.29,-.10);
  \path[draw=fwink!80,fill=#3] (0,.20) ellipse [x radius=.29cm,y radius=.10cm];
  \end{scope}%
}
\newcommand{\fwNetwork}[3]{%
  \begin{scope}[shift={(#1,#2)},xscale=.82,yscale=.964706,line width=.50pt]
  \foreach \a in {-.20,.20}{\foreach \b in {-.30,0,.30}{
    \draw[fwline!75] (-.40,\a) -- (0,\b);
    \draw[fwline!75] (0,\b) -- (.40,\a);
  }}
  \foreach \a in {-.20,.20}{
    \filldraw[draw=#3,fill=white] (-.40,\a) circle [radius=.068cm];
    \filldraw[draw=#3,fill=#3!28] (.40,\a) circle [radius=.068cm];
  }
  \foreach \b in {-.30,0,.30}{
    \filldraw[draw=#3,fill=#3!16] (0,\b) circle [radius=.068cm];
  }
  \end{scope}%
}

\resizebox{\linewidth}{!}{%
\begin{tikzpicture}[x=1cm,y=1cm,yscale=.85,
  font=\sffamily\fontsize{8}{9.5}\selectfont,text=fwink,
  card/.style={draw=fwline!80,fill=white,rounded corners=2.5pt,line width=.5pt},
  label/.style={align=center,inner sep=0pt,outer sep=0pt},
  title/.style={label,font=\sffamily\bfseries\fontsize{8}{9.5}\selectfont},
  detail/.style={label,font=\sffamily\fontsize{7}{8.5}\selectfont},
  flow/.style={-{Latex[length=1.6mm,width=1.15mm]},draw=fwline,line width=.65pt}]

\fill[fwsky,rounded corners=4pt] (0,0) rectangle (6.88,5.00);
\fill[fwmint,rounded corners=4pt] (7.10,0) rectangle (13.98,5.00);
\node[anchor=west,inner sep=0pt,font=\sffamily\bfseries\fontsize{8.5}{10}\selectfont,text=fwblue]
  at (.22,4.60) {1\quad SSL-based pseudo-labeling};
\node[anchor=west,inner sep=0pt,font=\sffamily\bfseries\fontsize{8.5}{10}\selectfont,text=fwteal]
  at (7.32,4.60) {2\quad Scoring and coreset selection};

\draw[card] (.25,2.65) rectangle (1.97,4.10);
\fwData{1.11}{3.72}{fwink!85}{white}
\node[title,text width=1.68cm] at (1.11,3.20) {Labeled};
\node[detail,text width=1.60cm] at (1.11,2.86) {$\mathcal D_L$ (GT)};
\draw[card] (.25,.85) rectangle (1.97,2.30);
\fwData{1.11}{1.92}{white}{fwline}
\node[title,text width=1.68cm] at (1.11,1.40) {Unlabeled};
\node[detail,text width=1.60cm] at (1.11,1.06) {$\mathcal D_U$};

\draw[card,draw=fwblue] (2.49,1.05) rectangle (4.12,4.10);
\fwNetwork{3.305}{3.55}{fwblue}
\node[title,text width=1.40cm] at (3.305,2.70) {SSL\\model};
\node[detail,text width=1.45cm] at (3.305,1.65) {FixMatch};
\draw[draw=fwline,line width=.65pt] (1.97,3.38) -- (2.23,3.38) -- (2.23,1.58) -- (1.97,1.58);
\draw[flow] (2.23,2.55) -- (2.49,2.55);

\draw[card,draw=fwblue] (4.67,1.05) rectangle (6.58,4.10);
\fwData{5.83}{3.61}{fwblue!25}{fwblue}
\fwData{5.42}{3.46}{fwink!85}{white}
\node[title,text width=1.65cm] at (5.625,2.70) {Training\\pool};
\node[detail,text width=1.65cm] at (5.625,1.85) {GT: $\mathcal D_L$};
\node[detail,text width=1.65cm] at (5.625,1.36) {PL: $\mathcal D_U$};
\draw[flow] (4.12,2.55) -- (4.67,2.55);
\node[detail,text=fwblue,text width=6.34cm] at (3.44,.35)
  {GT: ground truth\quad PL: pseudo-label};

\draw[card,draw=fwteal] (7.32,1.05) rectangle (9.00,4.10);
\fwNetwork{8.16}{3.55}{fwteal}
\node[title,text width=1.50cm] at (8.16,2.75) {Scoring\\model};
\draw[fwline,line width=.35pt] (7.56,1.55) -- (8.76,1.55);
\draw[fwline,line width=.35pt] (7.56,1.55) -- (7.56,2.05);
\draw[fwblue,line width=.7pt] plot coordinates
  {(7.60,1.60) (7.77,1.66) (7.94,1.63) (8.11,1.81) (8.29,1.84) (8.47,1.95) (8.71,1.99)};
\draw[fwteal,line width=.7pt] plot coordinates
  {(7.60,1.59) (7.77,1.75) (7.94,1.69) (8.11,1.79) (8.29,1.67) (8.47,1.82) (8.71,1.75)};
\node[detail,text width=1.50cm] at (8.16,1.24) {Dynamics};
\node[detail,text=fwteal,text width=2.20cm] at (8.28,.52) {Trained from\\scratch};
\draw[flow] (6.58,2.55) -- (7.32,2.55);

\draw[card] (9.46,2.85) rectangle (11.82,4.10);
\node[title,text width=2.12cm] at (10.64,3.87) {Semi-AUM};
\node[detail,text width=1.72cm] at (10.39,3.25) {Double-end\\pruning};
\draw[card] (9.46,1.05) rectangle (11.82,2.30);
\node[title,text width=2.12cm] at (10.64,2.07) {Semi-DUAL};
\node[detail,text width=1.72cm] at (10.39,1.45) {Beta\\sampling};
\draw[draw=fwline,line width=.65pt] (9.00,2.55) -- (9.23,2.55);
\draw[flow] (9.23,2.55) |- (9.46,3.48);
\draw[flow] (9.23,2.55) |- (9.46,1.68);

\begin{scope}[shift={(11.30,2.98)},x=.38cm,y=.58cm]
\draw[fwline,line width=.35pt] (0,0) -- (1,0);
\foreach \x/\h in {.03/.30,.22/.60,.41/.91,.60/.73,.79/.41}
  {\fill[fwline!22] (\x,0) rectangle +(.14,\h);}
\foreach \x/\h in {.22/.60,.41/.91,.60/.73}
  {\fill[fwblue!65] (\x,0) rectangle +(.14,\h);}
\draw[fwink,densely dashed,line width=.4pt] (.19,0) -- (.19,1.02);
\draw[fwink,densely dashed,line width=.4pt] (.77,0) -- (.77,1.02);
\end{scope}
\begin{scope}[shift={(11.30,1.18)},x=.38cm,y=.58cm]
\draw[fwline,line width=.35pt] (0,0) -- (1,0);
\foreach \x/\h in {.03/.27,.22/.48,.41/.67,.60/.56,.79/.37}
  {\fill[fwteal!28] (\x,0) rectangle +(.14,\h);}
\draw[fwteal,line width=.75pt] (0,.05)
  .. controls (.35,.02) and (.46,.95) .. (.66,.94)
  .. controls (.83,.91) and (.93,.38) .. (1,.08);
\end{scope}

\draw[card,draw=fwteal] (12.30,1.70) rectangle (13.85,3.55);
\fwData{13.00}{3.12}{fwteal!25}{fwteal}
\begin{scope}[shift={(13.25,2.95)},yscale=1.176471,scale=.78]
\filldraw[draw=fwteal,fill=white,line width=.55pt] (0,0) circle [radius=.16cm];
\draw[fwteal,line width=.8pt,line cap=round,line join=round]
  (-.09,0) -- (-.03,-.06) -- (.09,.07);
\end{scope}
\node[title,text width=1.45cm] at (13.075,2.29) {Selected\\coreset};
\draw[draw=fwline,line width=.65pt] (11.82,3.48) -- (12.06,3.48) -- (12.06,1.68) -- (11.82,1.68);
\draw[flow] (12.06,2.55) -- (12.30,2.55);

\draw[card,dashed] (12.15,.26) rectangle (13.85,1.16);
\begin{scope}[shift={(12.34,.74)},yscale=1.176471,rotate=-35,scale=.62]
\filldraw[draw=fwink,fill=fwblue!24,line width=.5pt]
  (-.075,-.17) rectangle (.075,.23);
\filldraw[draw=fwink,fill=white,line width=.5pt]
  (-.075,-.17) -- (0,-.32) -- (.075,-.17) -- cycle;
\fill[fwink] (-.026,-.27) -- (0,-.32) -- (.026,-.27) -- cycle;
\draw[fwink,line width=.4pt] (-.075,.15) -- (.075,.15);
\end{scope}
\node[detail,text width=1.25cm] at (13.16,.71) {GT for\\training};
\draw[flow,dashed] (13.075,1.70) -- (13.075,1.16);
\end{tikzpicture}%
}
\endgroup
\caption{\textbf{Overview of \SemiPrune.} An SSL model uses a small labeled subset to generate pseudo-labels, retaining initial GT labels. A separate scorer uses the combined pool for AUM-based double-end pruning or DUAL-based Beta sampling. Selected examples receive GT labels for downstream training and evaluation.}
\label{fig:framework}
\end{figure}

Figure~\ref{fig:framework} summarizes the two-stage pipeline of \SemiPrune{}:
SSL-based pseudo-label generation followed by supervised scoring and coreset selection.
We train an SSL model using a small randomly labeled subset and the remaining unlabeled data. Its pseudo-labels, together with the retained ground-truth labels, supervise a separate model trained from scratch. We use this model's training dynamics to score examples and select a coreset.

Following \ELFS{}~\citep{zheng2025elfs}, we use pseudo-labels to enable
supervised pruning without full-pool annotation.
Our key distinction is to generate these labels through SSL on the
target data, using a small labeled subset to guide target adaptation.

\subsection{Limitations of Deep-clustering-based Pseudo-labeling} 
To motivate this choice of pseudo-label generator, we examine
settings in which deep-clustering-based pseudo-labeling becomes
unreliable. The deep clustering approach adopted in \ELFS~\citep{zheng2025elfs} relies on features extracted from an ImageNet-pretrained foundation model~\citep{oquab2024dinov, adaloglou2023exploring}. Therefore, its pseudo-label quality critically depends on whether the pretrained feature space is well aligned with the semantic structure of the target dataset.

First, when the target dataset is weakly aligned with the pretraining data, deep clustering may fail to recover semantically meaningful clusters. Rather than capturing the task-relevant class structure, it can group examples according to superficial visual similarity. This issue becomes more evident in domain-specific or fine-grained datasets such as Food-101, whose discriminative factors differ substantially from those captured by generic ImageNet-pretrained representations.
\begin{figure}[t]
    \centering
    \resizebox{0.72\textwidth}{!}{%
    \begin{minipage}[b]{0.34\linewidth}
        \centering
        \includegraphics[width=\linewidth]{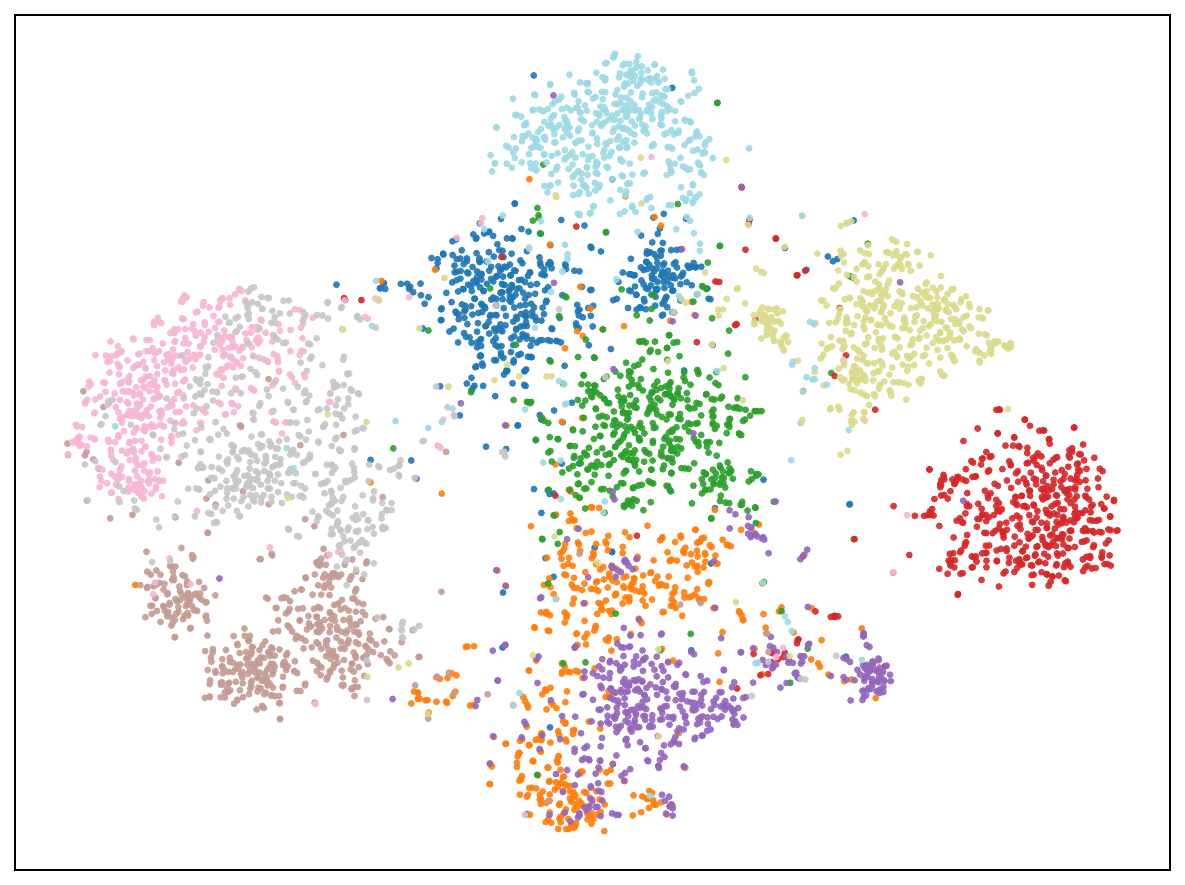}
    \end{minipage}
    \hfill
    \begin{minipage}[b]{0.34\linewidth}
        \centering
        \includegraphics[width=\linewidth]{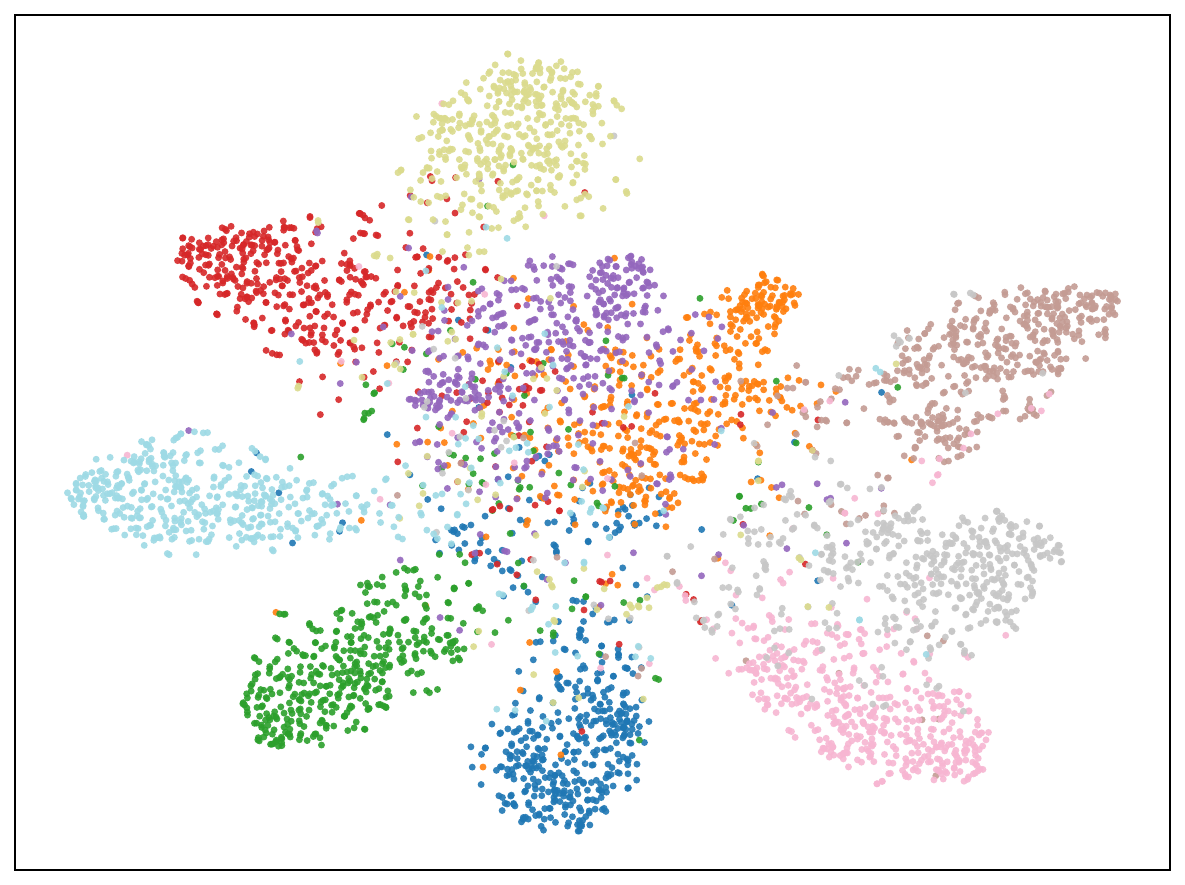}
    \end{minipage}
    }
    \caption{
        t-SNE visualization of Food-101 embeddings, colored by ground-truth labels. \textbf{(Left)} Embeddings from ImageNet-pretrained DINO. \textbf{(Right)} Embeddings from the SSL-trained model.
        }
    \label{fig:food_embedding}
\end{figure}

Figure~\ref{fig:food_embedding} compares Food-101 embeddings from
ImageNet-pretrained DINO~\citep{caron2021emerging} and a
FixMatch-trained model~\citep{sohn2020fixmatch}, using 500 examples
from each of 10 randomly selected classes.
The SSL-trained model exhibits clearer class separation despite using
only a 10\% initial label budget.
This observation is consistent with the higher pseudo-label accuracy
reported in Table~\ref{tab:pseudo_label_acc}.

Second, even when the target classes are reasonably aligned with
the pretraining distribution, image corruptions can obscure class
structure in pretrained representations, making clustering-based
pseudo-labels less reliable.
Figure~\ref{fig:cifar-c_embedding} in
Appendix~\ref{appendix:dc_ssl_embeddings_corruption} compares the representations
of the pretrained and SSL-trained models under corruption.

Third, class imbalance can distort feature-space clustering: broad majority-class regions may fragment, while sparse minority-class examples are absorbed into nearby clusters. The resulting pseudo-labels can misrepresent class identities and proportions, especially for underrepresented classes. SSL uses the labeled subset to guide predictions toward target classes and preserve class structure (Figure~\ref{fig:cifar-lt_embedding} and
Table~\ref{tab:pseudo_label_acc}).

\begin{table}[htbp]
\centering
\caption{Pseudo-label accuracy of deep clustering and SSL across diverse datasets. In dataset names, C denotes CIFAR and IN denotes ImageNet. For long-tailed datasets, we report balanced accuracy in parentheses. We adopt TEMI~\citep{adaloglou2023exploring} for deep clustering, and use a 10\% initial label budget to train FixMatch~\citep{sohn2020fixmatch} as an SSL algorithm.}

\label{tab:pseudo_label_acc}
\resizebox{0.85\textwidth}{!}{%
\begin{tabular}{lcc|cc|cc|ccc}
\toprule
\multirow{3}{*}{\makecell[l]{Method}} 
& \multicolumn{9}{c}{Pseudo-label ACC (\%)} \\
\cmidrule(lr){2-10}
& \multicolumn{2}{c|}{Domain-specific} 
& \multicolumn{2}{c|}{Image-corrupted} 
& \multicolumn{2}{c|}{Long-tailed}
& \multicolumn{3}{c}{Standard} \\
\cmidrule(lr){2-10}
& \hspace{.5em}Food & SUN & C-100 & Tiny-IN & C-100 & Caltech & C-10 & C-100 & IN-1K \\
\midrule
Deep-Clust.
& \hspace{.5em}42.8 & 38.1 & 40.3 & 40.9 & 21.9 (18.9) & 24.3 (12.7) & 95.1 & 66.6 & 57.3 \\
Semi-Sup.
& \hspace{.5em}61.7 & 49.2 & 65.4 & 58.0 & 65.4 (65.0) & 52.6 (44.5) & 95.4 & 74.1 & 64.0 \\
\bottomrule
\end{tabular}%
}
\end{table}

Across the evaluated settings, SSL-based pseudo-labeling achieves
higher accuracy than deep clustering
(Table~\ref{tab:pseudo_label_acc}), motivating its use as supervision
for downstream pruning.
These findings are consistent with theoretical work showing that
SSL can exploit the underlying data distribution under suitable
assumptions~\citep{wei2021theoretical}.

\subsection{SSL-based Pseudo-label Generation}
We implement this pseudo-label generation stage using
FixMatch~\citep{sohn2020fixmatch}, training for up to 100 epochs
on each dataset.\footnote{We shorten the original FixMatch schedule of about one million training iterations to reduce cost; Appendices~\ref{appendix:computation_time} and~\ref{appendix:end_to_end_cost} report pseudo-label generation time and end-to-end costs, respectively.}
Given a small labeled subset $\mathcal{D}_L$ with $(\mathbf{x}_i,y_i)\in\mathcal{D}_L$ and an unlabeled pool $\mathcal{D}_U$ with $|\mathcal{D}_L|\ll|\mathcal{D}_U|$,
FixMatch trains a model using the cross-entropy loss on $\mathcal{D}_L$ together with consistency regularization between weakly and strongly augmented views of unlabeled examples. Let $\phi$ denote the parameters of the FixMatch-trained model. For each unlabeled example $\mathbf{x}_i \in \mathcal{D}_U$, the model predicts a class distribution $p(\cdot \mid \mathbf{x}_i;\phi)$, and we assign the pseudo-label $\hat{y}_i = \arg\max_c p(c \mid \mathbf{x}_i;\phi)$. We then combine initially labeled examples, for which $\hat{y}_i = y_i$, with pseudo-labeled examples to form a pseudo-labeled training pool $\mathcal{D}_{\mathrm{train}}=\{(\mathbf{x}_i,\hat{y}_i)\}_{i=1}^{N}$, which is used in the subsequent score computation stage.

\subsection{Dataset Pruning with Pseudo-label-based Scores}
After constructing the pseudo-labeled training pool
$\mathcal{D}_{\mathrm{train}}=\{(\mathbf{x}_i,\hat{y}_i)\}_{i=1}^{N}$, we train a supervised model on $\mathcal{D}_{\mathrm{train}}$ from scratch and collect its prediction trajectories over $T$ epochs. Let $\theta_t$ denote the model parameters at epoch $t$ of supervised training, and let
$p(c\mid \mathbf{x}_i;\theta_t)$ denote the predicted probability of class $c$ for sample $\mathbf{x}_i$ under this model.
These dynamics are used to compute example-wise pruning scores with two existing criteria: \AUM{} with double-end pruning from~\citet{zheng2025elfs} and \DUAL{} with Beta sampling from~\citet{cho2025lightweight}. 
We refer to their pseudo-label-based variants as \textbf{\SemiAUM} and \textbf{\SemiDUAL}, respectively.

\subsubsection{\SemiAUM{} with Double-end Pruning}
We first follow the double-end pruning strategy introduced by \ELFS~\citep{zheng2025elfs} using the \AUM{} score~\citep{pleiss2020identifying}. 
For each example $(\mathbf{x}_i,\hat{y}_i)$, the margin at epoch $t$ is
$M(\mathbf{x}_i, \hat{y}_i;\theta_t) := 
p({\hat{y}_i} \mid \mathbf{x}_i;\theta_t)-\max_{c \neq \hat{y}_i} p(c\mid\mathbf{x}_i;\theta_t)$,
and the \AUM{} score is
$\mathrm{AUM}(\mathbf{x}_i,\hat{y}_i) := \tfrac{1}{T}\sum\nolimits_{t=1}^{T} M(\mathbf{x}_i,\hat{y}_i;\theta_t)$.
Low-AUM examples are hard to learn, whereas high-AUM examples are easily learned and may be less informative. 
Following \ELFS, we adopt a double-end pruning strategy to select examples from the middle of the \AUM{} distribution using an interval $[\alpha,\alpha+\Delta]$, where $\Delta$ is fixed by the target coreset size and $\alpha$ is tuned for each $\Delta$.

\subsubsection{\SemiDUAL{} with Beta Sampling}
We also consider \DUAL{} with Beta sampling, introduced by~\citet{cho2025lightweight}. It computes an example score from early-training dynamics by combining prediction difficulty and prediction uncertainty. For a window of length $J$ starting at epoch $k \in \{1,\ldots, T-J+1\}$, we write $p_{i,t} := p(\hat{y}_i \mid \mathbf{x}_i; \theta_t)$ and define the average probability assigned to the pseudo-label and its standard deviation as
\[
\bar{p}_{k,J} := \frac{1}{J}\sum_{j=0}^{J-1} p_{i,k+j}, \qquad
\sigma_{k,J} := \sqrt{\frac{1}{J-1}\sum_{j=0}^{J-1}\big(p_{i,k+j}-\bar{p}_{k,J}\big)^2}.
\]
Here, $1-\bar{p}_{k,J}$ measures prediction difficulty, while $\sigma_{k,J}$ measures prediction uncertainty within the window. 
The original \DUAL{} score within this window is calculated as
$\mathrm{DUAL}_{k,J}(\mathbf{x}_i, \hat y_i) := (1-\bar{p}_{k,J})\,\sigma_{k,J}$. 
To control the relative contribution of uncertainty under
pseudo-label supervision, we introduce an exponent $\gamma>0$:
$\mathrm{DUAL}_{k,J}^{(\gamma)}(\mathbf{x}_i,\hat y_i)
:= (1-\bar{p}_{k,J})\,\sigma_{k,J}^{\gamma}$,
where $\gamma=1$ recovers the original score.
The final example score averages these window-level scores:
\[
\mathrm{DUAL}^{(\gamma)}(\mathbf{x}_i,\hat y_i)
:= \frac{1}{T-J+1}\sum_{k=1}^{T-J+1}
\mathrm{DUAL}_{k,J}^{(\gamma)}(\mathbf{x}_i,\hat y_i).
\]

We then select the coreset following the original \DUALBeta{} procedure~\citep{cho2025lightweight}
to adjust the sampling distribution according to the pruning ratio.
Let $r \in [0,1]$ denote the pruning ratio and $\mu_D$ be the prediction mean of an example with a high DUAL score.
For each $r$, the Beta parameters are defined as
\begin{equation}
    \beta_r = C(1-\mu_D)(1-r^{c_D}), 
    \qquad 
    \alpha_r = C - \beta_r,
\end{equation}
where $C>0$ is a concentration constant and $c_D \ge 1$ controls the evolution rate of the distribution. Following the original DUAL setting, we keep $C$ fixed at $16$ and do not tune it. 
Because the mean of $\mathrm{Beta}(\alpha_r,\beta_r)$ is $\alpha_r/(\alpha_r+\beta_r)$, this choice shifts the sampling mean from $\mu_D$ at small pruning ratios toward $1$ at large pruning ratios.
Therefore, as pruning becomes more aggressive, the sampler increasingly favors easier examples.

\subsubsection{Hyperparameter Search with Pseudo-labeled Validation Set}
Following \ELFS~\citep{zheng2025elfs}, we use a 10\% held-out validation split for hyperparameter search, where the validation labels are given by the generated pseudo-labels. 
For \AUM{} with double-end pruning, we tune the cutoff ratio used to remove hard examples. 
For \DUAL{} with Beta sampling, we fix the window length to $J=10$ and tune the score-computation epoch $T$ and the dataset-specific parameter $c_D$. 
We use $\gamma=1$ except on CIFAR-10, CIFAR-100, and ImageNet, where it is tuned (Appendix~\ref{appendix:exp_details}); Table~\ref{tab:gamma_ablation} reports sensitivity to $\gamma$.
\section{Experiments}

\subsection{Experimental Setup}

\paragraph{Datasets.}
We primarily evaluate our framework under the challenging settings
motivated in Section~\ref{sec:methodology}: domain mismatch
(Food-101~\citep{bossard2014food} and SUN397~\citep{xiao2010sun}),
feature-space distortion (30\% corrupted CIFAR-100 and Tiny-ImageNet
following \citet{xia2023moderate}), and class imbalance
(Caltech-101~\citep{fei2004learning} and long-tailed CIFAR-100 with an
imbalance factor of 0.1).
We additionally evaluate on standard benchmarks, including
CIFAR-10/100~\citep{krizhevsky09learning} and
ImageNet-1K~\citep{deng2009imagenet}.

\paragraph{Baselines.}
We compare against \Random, \Prototypicality~\citep{sorscher2022beyond},
\ZCore~\citep{griffin2026zero}, \ScoreExtrap~\citep{schmidt2026effective},
and two \ELFS{} variants: the default with DINO features and a
self-trained SwAV encoder on the target data~\citep{zheng2025elfs}.
We evaluate two variants of our method:
\textbf{\SemiAUMCutoff} and \textbf{\SemiDUALBeta}.
We report the original supervised \DUALBeta~\citep{cho2025lightweight}
as a representative fully supervised method
(denoted as \FullySup{} in all tables).
Across all results, we highlight the best and second-best results among
methods that use either no labels or only a small fraction of labeled
data, excluding the fully supervised baseline from this ranking.
The best result is shown in \textbf{bold}, and the second-best is
\underline{underlined}.
Full experimental and baseline details are provided in
Appendix~\ref{appendix:exp_details}.

\paragraph{Evaluation protocol.}
Unless otherwise stated, the 10\% label budget refers to the initial
annotations used for SSL-based pseudo-label generation and coreset
selection. For downstream evaluation, the selected coreset is annotated
with ground-truth labels and a model is trained on it from scratch.
Thus, this initial budget does not include all annotations required for
final coreset training.

\subsection{Coreset Performance across Diverse Settings}
\label{sec:baseline_comparison}

We evaluate \SemiPrune{} under the three challenging data regimes that
motivate our framework: domain mismatch, image corruption, and class
imbalance.
We then consider standard benchmarks, where pretrained representations
are expected to provide a stronger basis for feature-based label-free
pruning methods.

\subsubsection{Domain Mismatch}

\begin{table}[!htbp]
\centering
\caption{Coreset performance on Food-101 and SUN397.
We report the average test accuracy over five runs.
The full-dataset test accuracies are 80.3\% and 61.4\%, respectively.}
\renewcommand{\arraystretch}{1.15}

\resizebox{0.9\textwidth}{!}{%
\begin{tabular}{llccccc|ccccc}
\toprule
\multirow{2}{*}{Label Usage} & \multirow{2}{*}{Pruning Rate}
& \multicolumn{5}{c|}{Food-101}
& \multicolumn{5}{c}{SUN397} \\
\cmidrule(lr){3-7} \cmidrule(lr){8-12}
& & 30\% & 50\% & 70\% & 80\% & 90\%
  & 30\% & 50\% & 70\% & 80\% & 90\% \\
\midrule

With Labels & \FullySup{}
& 80.2 & 77.5 & 71.3 & 67.2 & 55.3 
& 60.7 & 56.1 & 50.0 & 49.3 & 40.4 \\
\midrule

\multirow{5}{*}{Without Labels}
& \Random{}
& 76.6 & 72.9 & 63.9 & 55.5 & 37.7 
& 57.5 & 52.9 & 42.8 & 41.0 & 30.4 \\

& \Prototypicality{}
& 72.3 & 66.9 & 56.6 & 47.1 & 28.9 
& 59.1 & 52.9 & 39.6 & 33.9 & 21.0 \\

& \ELFSDINO{}
& 77.2 & 73.3 & 65.6 & 61.1 & 49.2
& \uline{60.6} & \textbf{56.1} & 46.7 & \uline{43.7} & 33.7 \\

& \ELFSSelfEnc{}
& 73.7 & 62.8 & 58.6 & 52.9 & 40.0 
& 59.5 & 55.1 & 45.2 & 41.9 & 31.3 \\

& \ZCore{}
& 76.7 & 72.1 & 63.4 & 56.3 & 37.8
& 59.4 & 55.0 & 44.7 & 42.0 & 29.9 \\

\midrule

\multirow{3}{*}{With 10\% Labels}
& \ScoreExtrap{}
& 76.3 & 71.8 & 63.2 & 57.6 & 42.3
& 58.7 & 53.6 & 43.9 & 39.9 & 30.0 \\

& \mainhl\SemiAUMCutoff{}
& \mainhl\textbf{77.9} & \mainhl\textbf{73.9} & \mainhl\textbf{67.2} & \mainhl\textbf{63.1} & \mainhl\uline{51.2}
& \mainhl\textbf{60.7} & \mainhl55.8 & \mainhl\textbf{47.7} & \mainhl\textbf{44.1} & \mainhl\textbf{35.0} \\

& \mainhl\SemiDUALBeta{}
& \mainhl\textbf{77.9} & \mainhl\uline{73.4} & \mainhl\uline{66.3} & \mainhl\uline{62.2} & \mainhl\textbf{51.8} 
& \mainhl60.2 & \mainhl\textbf{56.1} & \mainhl\uline{47.3} & \mainhl43.0 & \mainhl\uline{33.9} \\
\bottomrule

\end{tabular}%
}
\label{tab:fine_food_sun}
\end{table}

We first consider datasets whose visual characteristics differ
substantially from ImageNet-1K, the dataset used to pretrain many
off-the-shelf feature extractors.
Under such domain mismatch, the effectiveness of \ELFSDINO{} can be
limited because its ImageNet-pretrained representation may not preserve
the intrinsic geometry of the target dataset.
As illustrated in Figure~\ref{fig:food_embedding}, semantically meaningful
groups are not clearly separated, resulting in noisy pseudo-label
assignments.
These errors can in turn distort the training dynamics used for
difficulty estimation and degrade coreset performance.

Using only a 10\% labeled initial subset, our method generates
target-adaptive pseudo-labels and consistently achieves the best
coreset performance among all methods except the fully supervised
reference.
In contrast, \ScoreExtrap{} yields substantially lower coreset
performance despite using the same labeled subset.
These results support the use of limited target-class supervision for
generating pseudo-labels used in full-pool scoring.

\ELFSSelfEnc{} also underperforms \ELFSDINO{} and sometimes random
selection, indicating that self-supervised adaptation to the target
data alone does not necessarily yield representations suitable for
coreset selection.
Other label-free methods, such as \Prototypicality{} and \ZCore, also
depend heavily on pretrained representations and can perform below
random selection when those representations are poorly aligned with
the target data.

\subsubsection{Image Corruption}

\begin{table}[ht]
\centering
\caption{Coreset performance on 30\% corrupted CIFAR-100 and Tiny-ImageNet. We report the average test accuracy over five runs. The full-dataset test accuracies are 73.7\% and 44.3\%, respectively.}
\renewcommand{\arraystretch}{1.15}

\resizebox{0.9\textwidth}{!}{%
\begin{tabular}{llccccc|ccccc}
\toprule
\multirow{2}{*}{Label Usage} & \multirow{2}{*}{Pruning Rate}

& \multicolumn{5}{c|}{CIFAR-100-C}
& \multicolumn{5}{c}{Tiny-ImageNet-C} \\
\cmidrule(lr){3-7} \cmidrule(lr){8-12} 
& & 30\% & 50\% & 70\% & 80\% & 90\%
  & 30\% & 50\% & 70\% & 80\% & 90\% \\
\midrule

With Labels & \FullySup{}
& 73.6 & 71.1 & 65.2 & 61.1 & 52.6
& 43.7 & 41.3 & 35.5 & 31.4 & 24.2 \\
\midrule

\multirow{5}{*}{Without Labels}
& \Random{}
& 69.8 & 65.4 & 56.7 & 49.7 & 35.8 
& 38.9 & 34.4 & 27.1 & 22.6 & 17.0 \\

& \Prototypicality{}
& 72.5 & 68.1 & 58.3 & 50.4 & 35.7
& 39.3 & 30.2 & 18.4 & 12.2 & 6.1 \\

& \ELFSDINO{}
& \uline{73.1} & 68.7 & 61.1 & 56.6 & 44.9
& 40.9 & 37.4 & 32.5 & 29.4 & 22.7 \\

& \ELFSSelfEnc{}
& 71.5 & 68.0 & 60.2 & 54.1 & 38.6 
& 40.2 & 35.5 & 29.0 & 25.6 & 19.1 \\

& \ZCore{}
& 68.4 & 65.7 & 57.2 & 54.5 & 41.3
& 37.8 & 33.6 & 28.4 & 25.1 & 18.7 \\

\midrule
\multirow{3}{*}{With 10\% Labels}
& \ScoreExtrap{}
& 67.5 & 58.6 & 36.6 & 25.0 & 12.9
& \uline{41.3} & 31.7 & 32.1 & 28.3 & 23.0 \\

& \mainhl\SemiAUMCutoff{}
& \mainhl\uline{73.1} & \mainhl\uline{70.0} & \mainhl\textbf{63.9} & \mainhl\textbf{60.2} & \mainhl\textbf{52.0}
& \mainhl 40.9 & \mainhl\uline{37.7} & \mainhl\uline{33.4} & \mainhl\textbf{30.8} & \mainhl\textbf{24.1} \\

& \mainhl\SemiDUALBeta{}
& \mainhl\textbf{73.3} & \mainhl\textbf{70.4} & \mainhl\uline{63.8} & \mainhl\uline{58.6} & \mainhl\uline{47.5}
& \mainhl\textbf{42.0} & \mainhl\textbf{38.5} & \mainhl\textbf{33.8} & \mainhl\uline{30.1} & \mainhl\uline{24.0} \\

\bottomrule
\end{tabular}%
}
\label{tab:corrupt}
\end{table}

We next evaluate our framework on corrupted datasets, where visual
corruptions distort the representation space and make coreset
selection particularly challenging for methods that rely on fixed
features.
Corruption-induced artifacts can alter the geometry of examples in
feature space, making it difficult for representation-based methods to
preserve the underlying semantic class structure.

By learning directly from the corrupted target dataset with limited
class supervision, SSL produces more accurate pseudo-labels than deep
clustering (Table~\ref{tab:pseudo_label_acc});
Appendix~\ref{appendix:ps_distribution_corruption} further compares their
class-wise distributions.
Using these pseudo-labels for scoring yields stronger coreset
performance, with particularly large gains over \ELFSDINO{} at high
pruning ratios on CIFAR-100-C.

\ScoreExtrap{} faces a different limitation in this setting.
It trains a scoring model only on the initial labeled subset and
propagates the resulting scores to unlabeled examples through a
feature-based graph.
When corruption makes neighborhood relationships in the feature space
unreliable, this propagation can produce inaccurate importance
estimates for the unlabeled pool.

\subsubsection{Long-Tailed Distribution}

\begin{table}[!ht]
\centering
\caption{Coreset performance on CIFAR-100-LT with imbalance factor 0.1 and Caltech-101.
For CIFAR-100, we use the original balanced test set. Five runs are averaged.
The full-dataset test accuracies are 62.9\% and 78.6\%, respectively.}
\label{tab:cifar100lt_caltech}
\resizebox{0.9\textwidth}{!}{%
\begin{tabular}{llccccc|ccccc}
\toprule
\multirow{2}{*}{Label Usage} & \multirow{2}{*}{Pruning Rate}
& \multicolumn{5}{c|}{CIFAR-100-LT}
& \multicolumn{5}{c}{Caltech-101} \\
\cmidrule(lr){3-7} \cmidrule(lr){8-12}
& & 30\% & 50\% & 70\% & 80\% & 90\%
  & 30\% & 50\% & 70\% & 80\% & 90\% \\
\midrule

With Labels & \FullySup{}
& 58.1 & 54.9 & 43.5 & 35.9 & 27.1
& 79.2 & 72.4 & 59.7 & 54.9 & 42.7 \\
\midrule

\multirow{5}{*}{Without Labels}
& \Random{}
& 32.9 & 18.8 & 8.3 & 5.4 & 3.2
& 75.9 & 65.1 & 53.5 & 50.5 & 40.3 \\

& \Prototypicality{}
& 54.7 & 50.5 & 33.2 & 24.4 & 16.7
& 76.0 & 64.9 & 35.0 & 23.8 & 12.0 \\

& \ELFSDINO{}
& 49.5 & 49.1 & \uline{43.7} & 34.2 & 25.1
& 75.7 & 68.1 & 53.8 & 47.0 & 40.0 \\

& \ELFSSelfEnc{}
& \uline{56.0} & 51.3 & 40.7 & 29.6 & 20.4 
& 75.6 & 66.2 & 54.0 & 47.1 & 37.2 \\

& \ZCore{}
& 55.5 & \uline{51.5} & 38.2 & 32.0 & 19.8 
& 71.5 & 65.8 & 56.3 & 51.3 & 38.5 \\
\midrule

\multirow{3}{*}{With 10\% Labels}
& \ScoreExtrap{}
& 55.3 & 48.5 & 36.0 & 28.6 & 20.4
& 73.2 & 66.6 & 56.1 & 49.1 & 38.8 \\

& \mainhl\SemiAUMCutoff{}
& \mainhl50.4 & \mainhl46.5 & \mainhl\textbf{43.8} & \mainhl\uline{34.4} & \mainhl\uline{27.0}
& \mainhl\uline{78.4} & \mainhl\uline{70.6} & \mainhl\uline{58.5} & \mainhl\uline{53.1} & \mainhl\uline{45.7} \\

& \mainhl\SemiDUALBeta{}
& \mainhl\textbf{58.3} & \mainhl\textbf{53.6} & \mainhl41.9 & \mainhl\textbf{35.1} & \mainhl\textbf{27.5} 
& \mainhl\textbf{79.0} & \mainhl\textbf{70.7} & \mainhl\textbf{58.8} & \mainhl\textbf{53.9} & \mainhl\textbf{46.8} \\

\bottomrule
\end{tabular}%
}
\end{table}

We next evaluate under class imbalance using CIFAR-100-LT with an
imbalance factor of 0.1 and the naturally imbalanced Caltech-101
dataset.
Unlike image corruption, class imbalance does not directly alter the
pretrained feature extractor, but instead changes the sample
distribution within the embedding space.

Under this imbalance, deep clustering may produce pseudo-labels whose
class distribution deviates from the true long-tailed distribution.
For example, clustering can split broad majority-class regions into
multiple clusters while merging nearby minority-class regions.
We further compare the class-wise distributions of generated
pseudo-labels in Appendix~\ref{appendix:ps_distribution_longtailed}, which
supports this distributional mismatch.
In contrast, SSL-based pseudo-labeling uses a small labeled subset to
anchor predictions to the target class space, allowing the resulting
pseudo-labels to better reflect the underlying long-tailed
distribution.

The benefit of these pseudo-labels nevertheless depends on the scoring
criterion and pruning ratio.
On CIFAR-100-LT, \SemiAUMCutoff{} trails \ELFSDINO{} at 50\% pruning,
whereas \SemiDUALBeta{} outperforms it at most evaluated pruning ratios.
\SemiDUALBeta{} also exceeds the fully supervised \DUALBeta{} reference
at 30\% and 90\% pruning on CIFAR-100-LT and at 90\% pruning on
Caltech-101.
Additional CIFAR-100-LT results at an imbalance factor of 0.01 and scorer-choice discussion appear in Appendices~\ref{appendix:severe_imbalance} and~\ref{appendix:scorer_choice}, respectively.

\subsubsection{Standard Benchmarks}
\label{sec:standard_benchmarks}

\begin{table}[!htbp]
\centering
\caption{Coreset performance on standard benchmarks.
The full-dataset test accuracies are 95.3\%, 78.9\%, and 73.6\%, respectively. 
We report the average test accuracy over five runs for CIFAR datasets and the test accuracy from a single run for ImageNet.}
\renewcommand{\arraystretch}{1.25}
\resizebox{\linewidth}{!}{%
\begin{tabular}{llccccc|ccccc|ccccc}
\toprule
\multirow{2}{*}{Label Usage} & \multirow{2}{*}{Pruning Rate}
& \multicolumn{5}{c|}{CIFAR-10}
& \multicolumn{5}{c|}{CIFAR-100}
& \multicolumn{5}{c}{ImageNet-1K} \\
\cmidrule(lr){3-7} \cmidrule(lr){8-12} \cmidrule(lr){13-17}
& & 30\% & 50\% & 70\% & 80\% & 90\%
  & 30\% & 50\% & 70\% & 80\% & 90\%
  & 30\% & 50\% & 70\% & 80\% & 90\% \\
\midrule

With Labels & \FullySup{}
& 95.5 & 95.2 & 93.0 & 91.4 & 87.1
& 77.9 & 74.7 & 69.3 & 64.8 & 54.5 
& 73.3 & 72.3 & 69.4 & 66.5 & 60.0 \\
\midrule

\multirow{5}{*}{Without Labels}
& \Random{}
& 94.4 & 93.2 & 90.5 & 88.3 & 83.7
& 75.2 & 71.7 & 64.9 & 59.2 & 45.1 
& 72.2 & 70.3 & 66.7 & 62.5 & 52.3 \\

& \Prototypicality{}
& 95.1 & 93.9 & 87.2 & 79.7 & 59.1
& 75.5 & 69.1 & 53.2 & 39.1 & 19.6 
& 70.9 & 60.8 & 54.6 & 41.9 & 30.6 \\

& \ELFSDINO{}
& 95.3 & \uline{95.1} & \uline{93.0} & \textbf{91.3} & \uline{86.9}
& \uline{77.3} & \textbf{73.8} & \uline{67.4} & 61.9 & 53.0
& \textbf{73.5} & \uline{71.8} & 67.2 & 63.4 & \uline{54.9} \\

& \ELFSSelfEnc{}
& 94.8 & 93.5 & 91.2 & 89.5 & 84.5
& 75.5 & 71.7 & 64.8 & 59.5 & 46.9 
& 73.2 & 71.4 & 66.8 & 62.7 & 53.4 \\

& \ZCore{}
& 94.6 & 93.5 & 91.0 & 89.1 & 84.2
& 76.0 & 72.9 & 65.9 & 61.9 & 52.1 
& 72.3 & 70.8 & 66.4 & 62.1 & 53.1 \\
\midrule

\multirow{3}{*}{With 10\% Labels}
& \ScoreExtrap{}
& 95.3 & 94.6 & 91.9 & 89.4 & 81.5
& 75.5 & 70.8 & 61.9 & 53.8 & 34.9 
& 68.9 & 65.2 & 59.1 & 55.2 & 53.1 \\

& \mainhl\SemiAUMCutoff{}
& \mainhl\textbf{95.4} & \mainhl\textbf{95.2} & \mainhl\textbf{93.1}& \mainhl\textbf{91.3} & \mainhl\textbf{87.2}
& \mainhl\textbf{77.4} & \mainhl73.5 & \mainhl\textbf{67.7} & \mainhl\textbf{63.1} & \mainhl\textbf{53.9}
& \mainhl\uline{73.3} & \mainhl\textbf{72.1} & \mainhl\uline{67.7} & \mainhl\uline{63.8} & \mainhl54.2 \\

& \mainhl\SemiDUALBeta{}
& \mainhl\textbf{95.4} & \mainhl95.0 & \mainhl92.7 & \mainhl90.9 & \mainhl86.2
& \mainhl77.1 & \mainhl\textbf{73.8} & \mainhl67.3 & \mainhl\uline{62.9} & \mainhl\uline{53.1} 
& \mainhl72.5 & \mainhl71.4 & \mainhl\textbf{67.8} & \mainhl\textbf{64.6} & \mainhl\textbf{57.8} \\
\bottomrule
\end{tabular}%
}
\label{tab:classic}
\end{table}

Finally, we evaluate on standard benchmarks, where ImageNet-pretrained
representations are expected to be more reliable and therefore provide
a more favorable setting for \ELFSDINO{}.
\SemiPrune{} remains competitive in these settings, with gains varying
across scoring criteria and pruning ratios.
For example, \SemiDUALBeta{} exceeds \ELFSDINO{} by
2.9 percentage points on ImageNet-1K at 90\% pruning.

\ELFSSelfEnc{} still performs poorly, indicating that a self-trained
encoder does not necessarily provide a reliable substitute for a
strong pretrained representation.
Under the same initial annotation budget, both \SemiPrune{} variants
outperform \ScoreExtrap{} at every evaluated pruning ratio on these
standard benchmarks.
These results show that SSL-based pseudo-labeling can effectively
exploit limited target labels even when pretrained representations
already provide a strong baseline.

\subsection{Analysis and Ablations}

\subsubsection{Effect of the Initial Label Budget}

\begin{figure}[htbp]
    \centering
    \includegraphics[width=\textwidth]{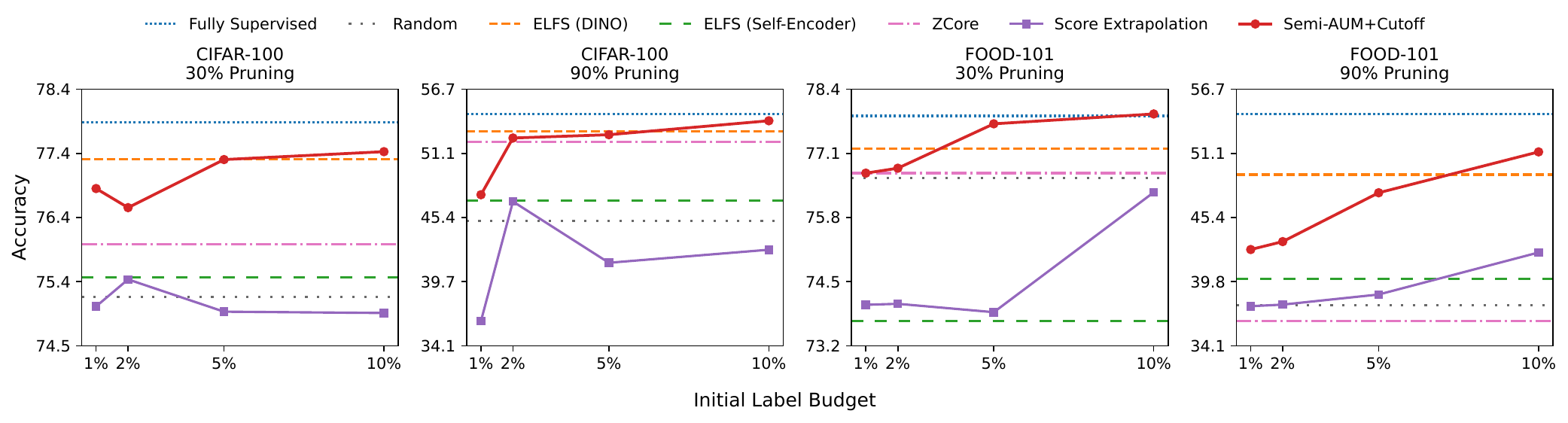}
    \caption{Effect of the initial label budget on coreset performance.
    We compare our method with \ScoreExtrap{} under identical initial
    label budgets and include label-free baselines for reference.}
    \label{fig:label_abl}
\end{figure}

We next examine the sensitivity of \SemiPrune{} to the amount of
target-class supervision.
Figure~\ref{fig:label_abl} compares \SemiAUMCutoff{} with \ScoreExtrap{}
under matched initial label budgets on CIFAR-100 and Food-101.
\SemiAUMCutoff{} consistently outperforms \ScoreExtrap{} across the
evaluated budgets and can exceed \ELFSDINO{} using only 5\% labeled
data.
It also outperforms \ELFSSelfEnc{} with a 1\% initial label budget.
Additional pruning ratios are reported in
Appendix~\ref{appendix:label_budget}.

These results support the motivation in Section~\ref{sec:methodology}: limited target-class supervision helps adapt scoring when pretrained features are poorly aligned with the target data. The comparison with \ELFSSelfEnc{} further suggests that this supervision can be more effective than self-supervised target adaptation alone in the evaluated settings.

Additional analyses of the SSL algorithm and pseudo-label generator architecture are provided in Appendices~\ref{appendix:ssl_generality} and~\ref{appendix:architecture}, respectively.

\subsubsection{Contribution of Training-Dynamics Scoring}

To isolate the contribution of training-dynamics scoring, we compare
\SemiPrune{} with an \textit{SSL Confidence} baseline that ranks
examples directly by the maximum confidence of the SSL pseudo-label
generator.
The baseline uses the same double-end selection and validation
procedure as \SemiAUMCutoff{}, thereby isolating the choice of scoring
signal.

\begin{table}[htbp]
\centering
\caption{Comparison of direct SSL-confidence ranking and training-dynamics scoring.
Values are mean $\pm$ sample standard deviation.}
\label{tab:ssl_confidence}
\vspace{0.4em}
\resizebox{0.9\linewidth}{!}{%
\renewcommand{\arraystretch}{1.15}
\begin{tabular}{llccccc}
\toprule
Dataset & Method & 30\% & 50\% & 70\% & 80\% & 90\% \\
\midrule

\multirow{3}{*}{Food-101}
 & SSL Confidence  & 77.2 $\pm$ 0.17 & 73.4 $\pm$ 0.18 & 66.3 $\pm$ 0.09 & 61.4 $\pm$ 0.20 & 50.0 $\pm$ 0.46 \\
 & \SemiAUM{}      & 77.9 $\pm$ 0.09 & 73.9 $\pm$ 0.09 & 67.2 $\pm$ 0.21 & 63.1 $\pm$ 0.27 & 51.2 $\pm$ 0.36 \\
 & \SemiDUAL{}     & 77.9 $\pm$ 0.16 & 73.4 $\pm$ 0.21 & 66.3 $\pm$ 0.14 & 62.2 $\pm$ 0.22 & 51.8 $\pm$ 0.32 \\
\midrule

\multirow{3}{*}{CIFAR-100-C}
 & SSL Confidence  & 69.5 $\pm$ 0.52 & 65.4 $\pm$ 0.52 & 56.7 $\pm$ 1.52 & 50.2 $\pm$ 1.05 & 37.0 $\pm$ 0.88 \\
 & \SemiAUM{}      & 73.1 $\pm$ 0.19 & 70.0 $\pm$ 0.31 & 63.9 $\pm$ 0.59 & 60.2 $\pm$ 0.42 & 52.0 $\pm$ 0.31 \\
 & \SemiDUAL{}     & 73.3 $\pm$ 0.28 & 70.4 $\pm$ 0.39 & 63.8 $\pm$ 0.76 & 58.6 $\pm$ 0.68 & 47.5 $\pm$ 0.81 \\
\midrule

\multirow{3}{*}{CIFAR-100-LT}
 & SSL Confidence  & 55.5 $\pm$ 0.86 & 52.0 $\pm$ 0.59 & 40.5 $\pm$ 1.80 & 34.1 $\pm$ 0.96 & 22.3 $\pm$ 1.38 \\
 & \SemiAUM{}      & 50.4 $\pm$ 1.17 & 46.5 $\pm$ 0.58 & 43.8 $\pm$ 0.62 & 34.4 $\pm$ 0.82 & 27.0 $\pm$ 0.87 \\
 & \SemiDUAL{}     & 58.3 $\pm$ 0.94 & 53.6 $\pm$ 0.73 & 41.9 $\pm$ 0.79 & 35.1 $\pm$ 0.91 & 27.5 $\pm$ 0.96 \\
\midrule

\multirow{3}{*}{CIFAR-100}
 & SSL Confidence  & 75.8 $\pm$ 0.12 & 72.4 $\pm$ 0.60 & 65.8 $\pm$ 0.44 & 59.7 $\pm$ 0.45 & 47.9 $\pm$ 0.83 \\
 & \SemiAUM{}      & 77.4 $\pm$ 0.23 & 73.5 $\pm$ 0.10 & 67.7 $\pm$ 0.26 & 63.1 $\pm$ 0.21 & 53.9 $\pm$ 0.23 \\
 & \SemiDUAL{}     & 77.1 $\pm$ 0.18 & 73.8 $\pm$ 0.27 & 67.3 $\pm$ 0.24 & 62.9 $\pm$ 0.31 & 53.1 $\pm$ 0.49 \\
\bottomrule
\end{tabular}%
}
\end{table}

Across diverse datasets and pruning ratios, at least one
training-dynamics-based variant consistently outperforms direct
SSL-confidence ranking.

SSL confidence reflects the pseudo-label generator's confidence in its own predictions, whereas \SemiAUM{} and \SemiDUAL{} capture how
individual examples behave during the optimization of a separate
scoring model.
These results support training a separate scorer to capture signals beyond pseudo-label confidence.
Appendix~\ref{appendix:pl_error_effects} provides class-wise and controlled-noise analyses linking pseudo-label errors to score rankings, coreset selection, and downstream performance.

We additionally compare SSL-based and deep-clustering-based pseudo-labels
under Forgetting, EL2N, and AUM in
Appendix~\ref{appendix:scoring_criterion}.

\section{Conclusion and Limitations}
\label{sec:conclusion}
We propose \SemiPrune{}, an SSL-based framework that extends supervised
training-dynamics-based pruning to limited-label settings.
Using a small labeled subset, \SemiPrune{} generates target-adaptive
pseudo-labels that enable existing pruning criteria to score the full
data pool without full-pool annotation.
This target-specific supervision is particularly useful when pretrained
representations are poorly aligned with the target domain.
Our results demonstrate improved coreset performance over label-free
baselines while requiring substantially fewer annotations than fully
supervised pruning.
\SemiPrune{} also consistently outperforms \ScoreExtrap{} under the
same initial annotation budget, highlighting the effectiveness of SSL-based
pseudo-labeling for leveraging limited labels in dataset pruning.

The main limitations of \SemiPrune{} are its reliance on an initial
labeled subset and the additional computation required for SSL training
(Appendix~\ref{appendix:end_to_end_cost}).
An insufficient label budget or poor class coverage in the labeled subset can compromise pseudo-label quality, affecting score rankings and coreset selection (Appendix~\ref{appendix:pl_noise}).
Although our experiments demonstrate compatibility with both FixMatch and FlexMatch, broader evaluation across SSL methods and alternative proxy-label sources, including VLM-generated labels, remains future work
(Appendix~\ref{appendix:ssl_generality}).

\begin{ack}
This work was supported by two Institute of Information \& Communications Technology Planning \& Evaluation (IITP) grants (No.\ RS-2024-00457882 National AI Research Lab Project, 80\%; No.\ RS-2019-II190075 Artificial Intelligence Graduate School Program (KAIST), 20\%) funded by the Korean government (MSIT). We also thank Nota Inc. for supporting our participation in the conference.
\end{ack}

\bibliographystyle{plainnat}
\bibliography{references}
\clearpage
\appendix
\addtocontents{toc}{\protect\setcounter{tocdepth}{3}}
\begingroup
\renewcommand{\contentsname}{Appendix Contents}
\tableofcontents
\endgroup
\clearpage
\section{Additional Pseudo-Label Analysis}
\label{appendix:pseudo_label_analysis}
Beyond the domain-mismatch examples in the main text, we compare representations, pseudo-label quality, and assigned class distributions under image corruption and class imbalance. We then examine class-wise errors and their consequences for scoring and selection. Pseudo-label quality is measured against ground-truth labels using accuracy (ACC), adjusted random index (ARI), and normalized mutual information (NMI), reported in \%.

\subsection{Pseudo-Label Quality under Distribution Shift}
\label{appendix:pl_quality_shift}

\subsubsection{Image Corruption}
\label{appendix:dc_ssl_embeddings_corruption}
\label{appendix:ps_distribution_corruption}
\begin{figure}[htbp]
    \centering
    \resizebox{0.65\textwidth}{!}{%
    \begin{minipage}[b]{0.34\linewidth}
        \centering
        \includegraphics[width=\linewidth]{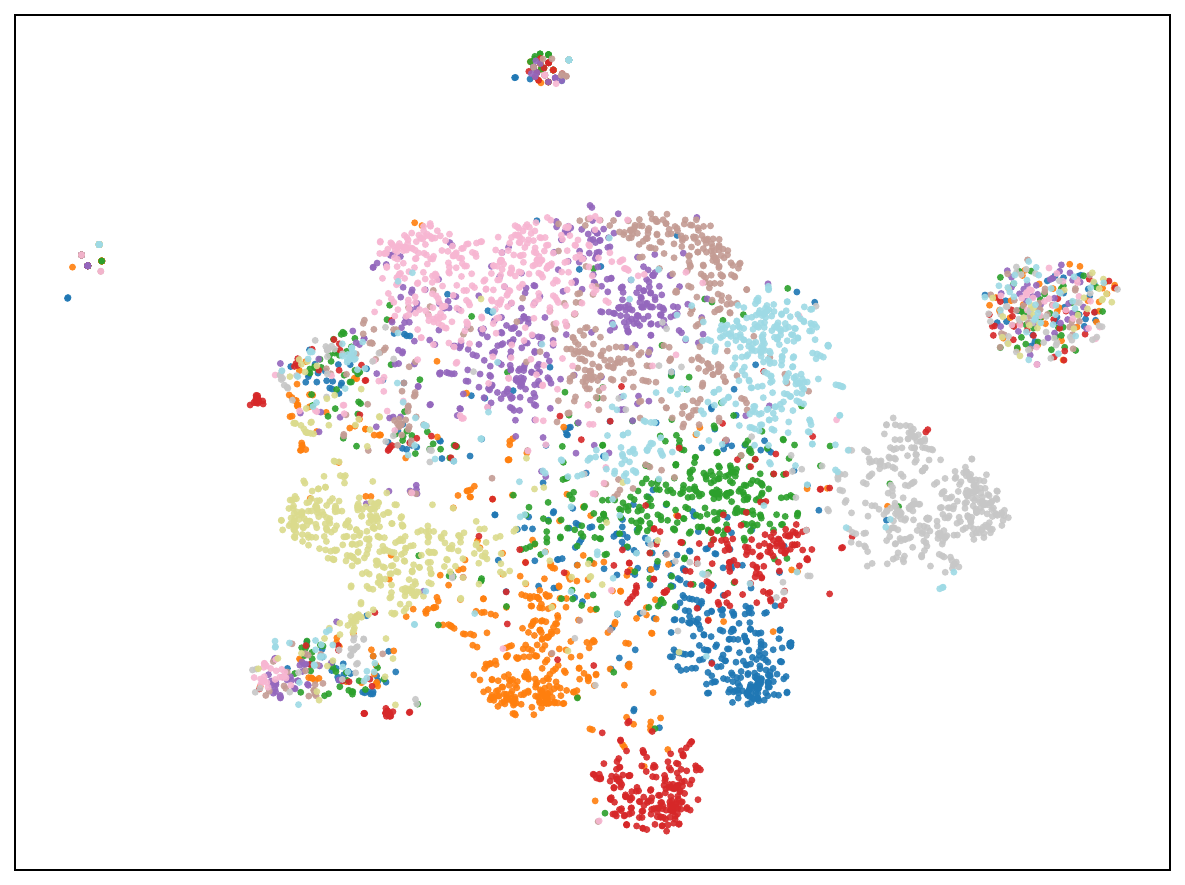}
    \end{minipage}
    \hfill
    \begin{minipage}[b]{0.34\linewidth}
        \centering
        \includegraphics[width=\linewidth]{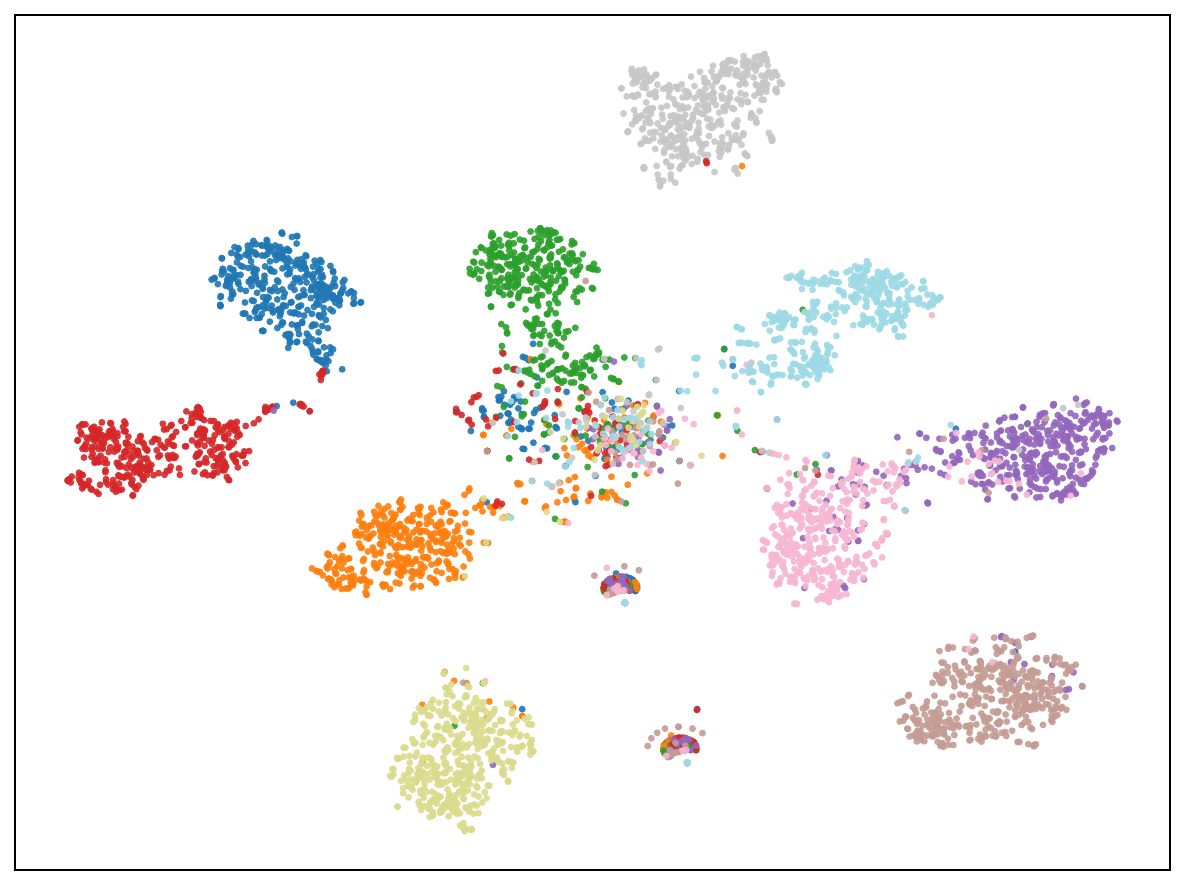}
    \end{minipage}
    }
    \caption{
        t-SNE visualization of corrupted CIFAR-100 embeddings, colored by ground-truth labels. \textbf{(Left)} Embeddings from ImageNet-pretrained DINO. \textbf{(Right)} Embeddings from the SSL-trained model.
        }
    \label{fig:cifar-c_embedding}
\end{figure}

The embeddings of CIFAR-100-C in Figure~\ref{fig:cifar-c_embedding} illustrate how corruption affects the representation space. Deep-clustering boundaries become less distinct than on clean data, whereas SSL better preserves separation between semantic classes.

Table~\ref{tab:pl_quality_corruption} quantifies pseudo-label quality on CIFAR-100-C and Tiny-ImageNet-C. SSL outperforms deep clustering on all three reported metrics for both datasets.

\begin{table}[htbp]
\centering
\caption{Pseudo-labeling quality on CIFAR-100-C and Tiny-ImageNet-C.}
\label{tab:pl_quality_corruption}
\renewcommand{\arraystretch}{1.0}
\resizebox{0.85\textwidth}{!}{%
\begin{tabular}{lccc|ccc}
\toprule
\multirow{2}{*}{Method} 
& \multicolumn{3}{c|}{CIFAR-100-C} 
& \multicolumn{3}{c}{Tiny-ImageNet-C} \\
\cmidrule(lr){2-4} \cmidrule(lr){5-7}
& ACC (\%) & ARI (\%) & NMI (\%) 
& ACC (\%) & ARI (\%) & NMI (\%) \\
\midrule
Deep Clustering
& 40.3 & 23.5 & 52.4
& 40.9 & 28.0 & 54.5 \\
Semi-Supervised
& 65.4 & 36.9 & 61.2
& 58.0 & 35.1 & 56.3 \\
\bottomrule
\end{tabular}%
}
\end{table}

\begin{figure}[htbp]
    \centering
    \begin{minipage}[t]{0.85\linewidth}
        \centering
        \includegraphics[width=\linewidth]{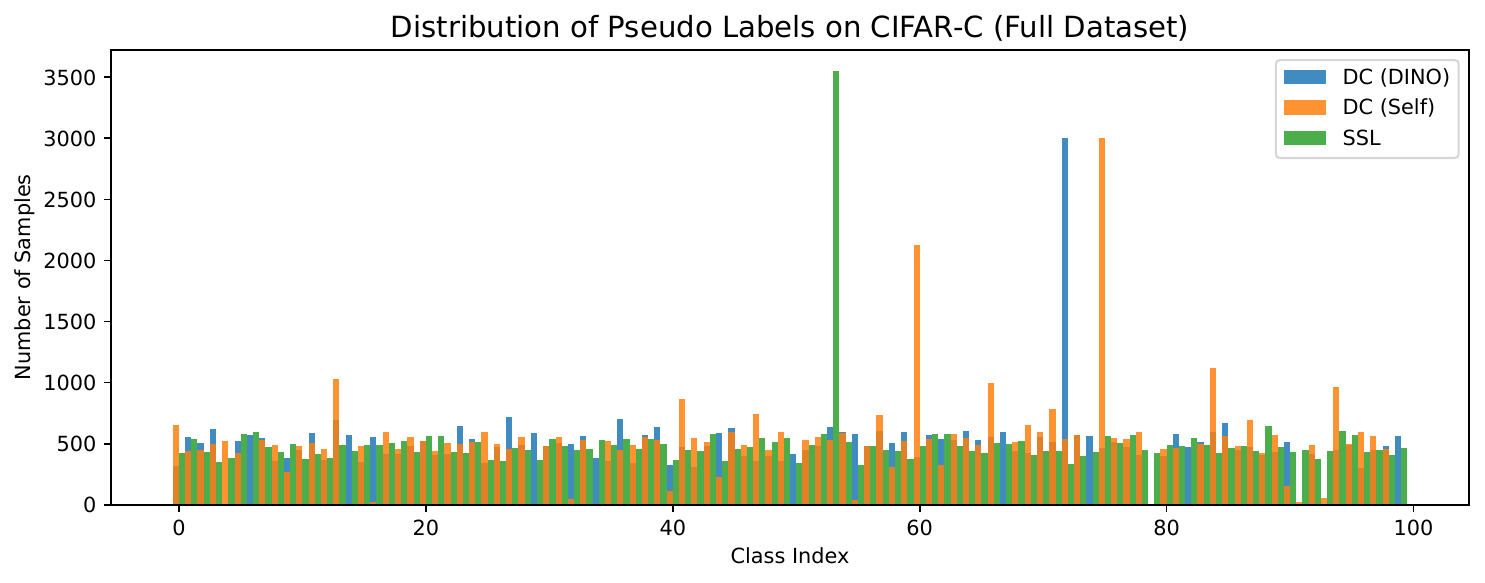}
    \end{minipage}
    \vspace{0.5em}
    \begin{minipage}[t]{0.85\linewidth}
        \centering
        \includegraphics[width=\linewidth]{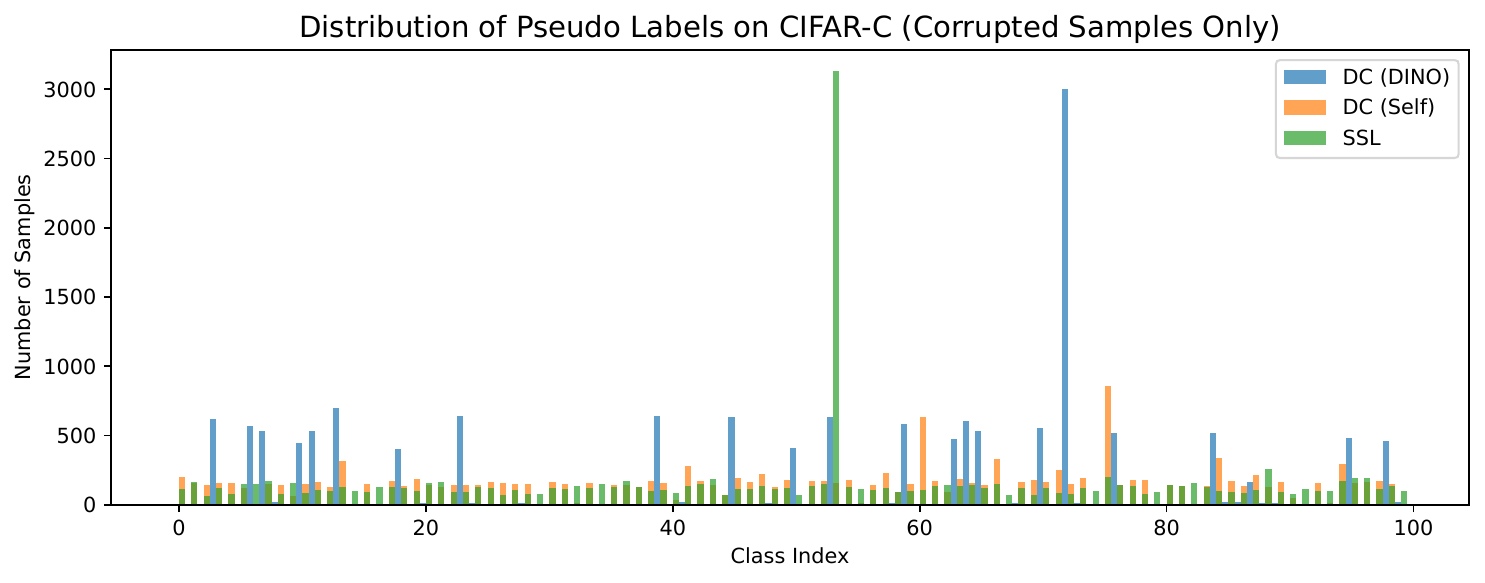}
    \end{minipage}
    \vspace{0.5em}
    \caption{Class-wise distribution of assigned pseudo-labels. \textbf{(Top)} Full dataset. \textbf{(Bottom)} Corrupted samples only. Deep-Clustering does not distribute pseudo-label assignments evenly across classes, leaving some classes with no assigned samples at all, whereas Semi-Supervised Learning yields a relatively more balanced distribution. For corrupted samples, both DC and SSL assign a substantial portion of samples to particular classes. However, SSL maintains a relatively broad distribution across classes, while DC shows a stronger tendency to concentrate assignments on only a few specific classes.}
    \label{fig:cifar-c_ps_dist}
\end{figure}

Figure~\ref{fig:cifar-c_ps_dist} complements these metrics with the class-wise distribution of assigned pseudo-labels, shown for both the full pool and the corrupted examples alone.

\FloatBarrier

\subsubsection{Long-Tailed Distributions}
\label{appendix:dc_ssl_embeddings_longtailed}
\label{appendix:ps_distribution_longtailed}
\begin{figure}[htbp]
    \centering
    \resizebox{0.65\textwidth}{!}{%
    \begin{minipage}[b]{0.34\linewidth}
        \centering
        \includegraphics[width=\linewidth]{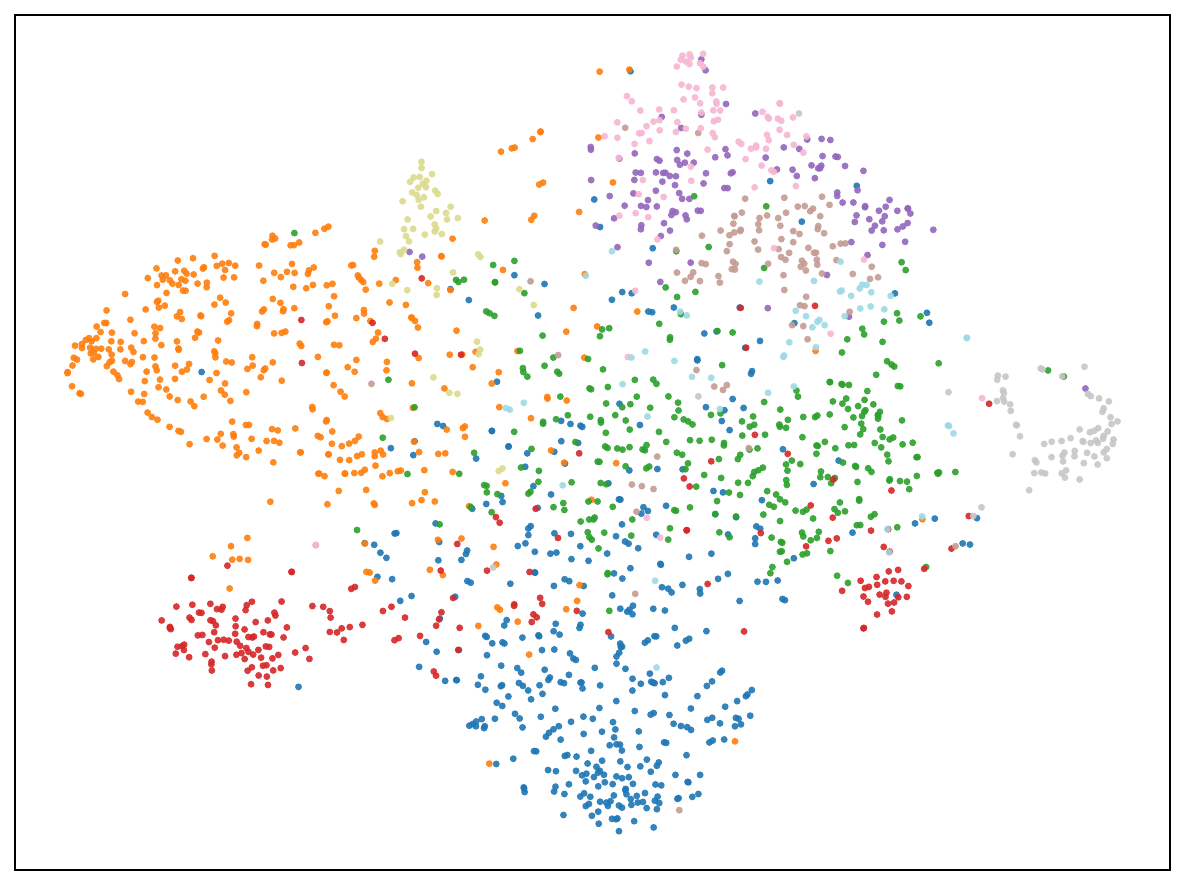}
    \end{minipage}
    \hfill
    \begin{minipage}[b]{0.34\linewidth}
        \centering
        \includegraphics[width=\linewidth]{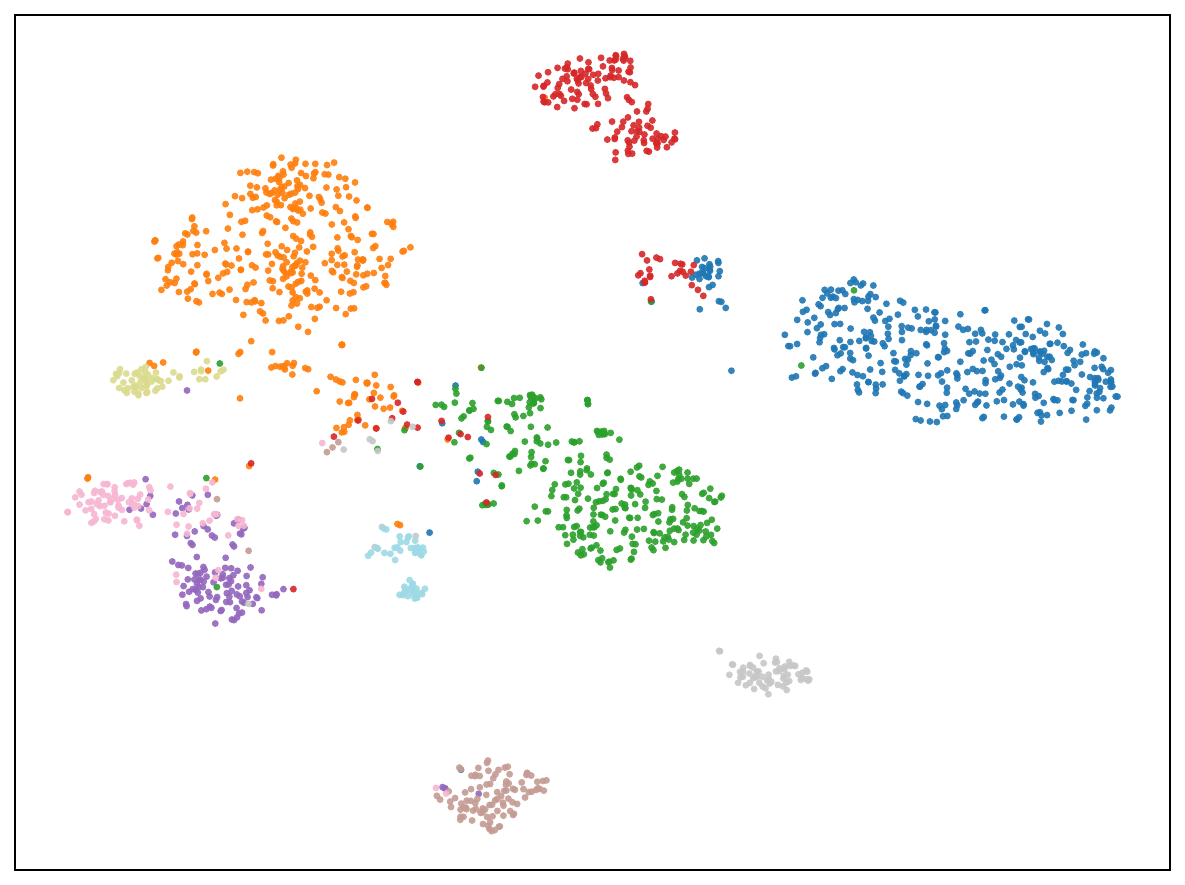}
    \end{minipage}
    }
    \caption{
        t-SNE visualization of long-tailed CIFAR-100 embeddings with an imbalance factor of 0.1, colored by ground-truth labels. \textbf{(Left)} Embeddings from ImageNet-pretrained DINO. \textbf{(Right)} Embeddings from the SSL-trained model.
        }
    \label{fig:cifar-lt_embedding}
\end{figure}

\begin{figure}[htbp]
    \centering
    \resizebox{0.65\textwidth}{!}{%
    \begin{minipage}[b]{0.34\linewidth}
        \centering
        \includegraphics[width=\linewidth]{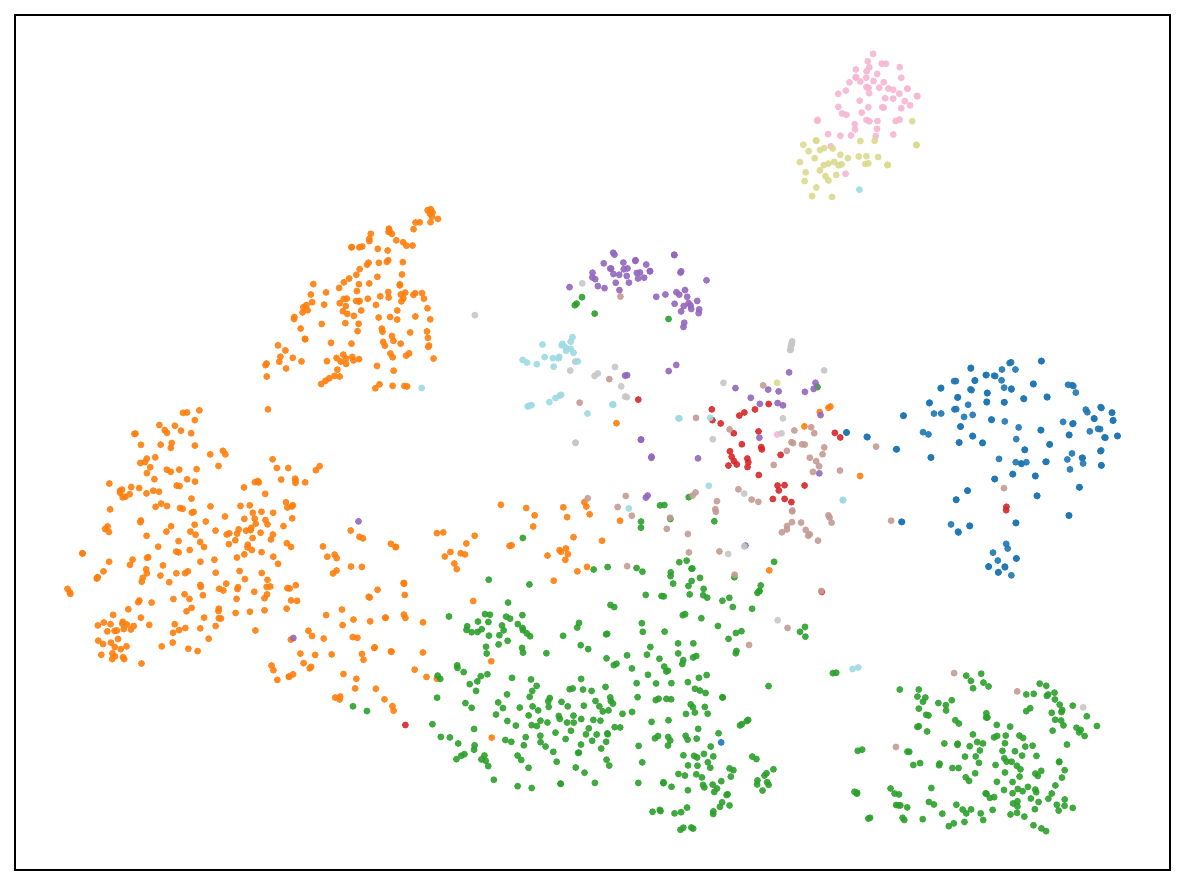}
    \end{minipage}
    \hfill
    \begin{minipage}[b]{0.34\linewidth}
        \centering
        \includegraphics[width=\linewidth]{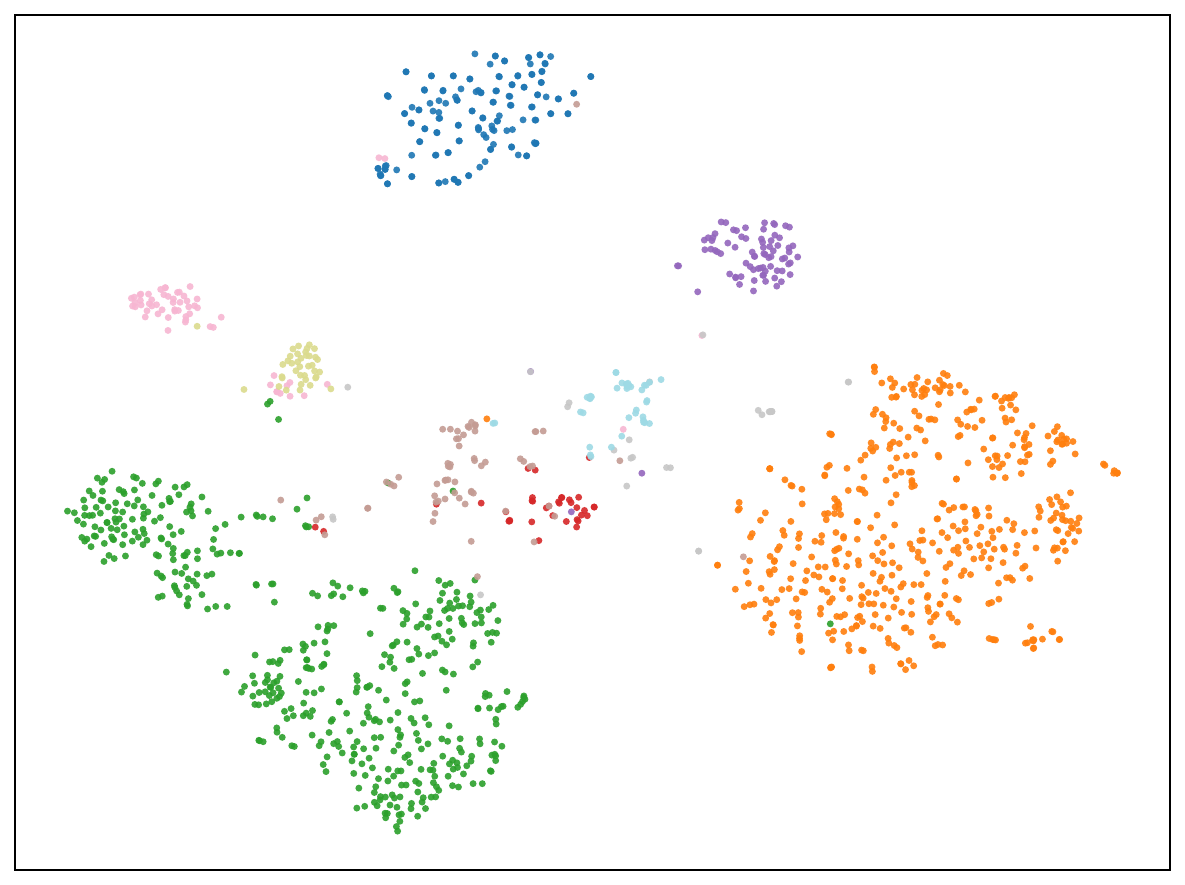}
    \end{minipage}
    }
    \caption{
        t-SNE visualization of Caltech-101 embeddings, colored by ground-truth labels. \textbf{(Left)} Embeddings from ImageNet-pretrained DINO. \textbf{(Right)} Embeddings from the SSL-trained model.
        }
    \label{fig:caltech_embedding}
\end{figure}

We next examine Caltech-101 and CIFAR-100-LT. Figures~\ref{fig:cifar-lt_embedding} and~\ref{fig:caltech_embedding} visualize the representations of CIFAR-100-LT with an imbalance factor of 0.1 and Caltech-101, respectively.

Table~\ref{tab:pl_quality_longtailed} reports pseudo-label quality on Caltech-101 and CIFAR-100-LT under two imbalance settings.

\begin{table}[htbp]
\centering
\caption{Pseudo-labeling quality on Caltech-101 and CIFAR-100-LT under two imbalance settings.}
\label{tab:pl_quality_longtailed}
\resizebox{\textwidth}{!}{%
\renewcommand{\arraystretch}{1.25}
\begin{tabular}{lccc|ccc|ccc}
\toprule
\multirow{2}{*}{Method}
& \multicolumn{3}{c|}{Caltech-101}
& \multicolumn{3}{c|}{CIFAR-100-LT, IF = 0.1}
& \multicolumn{3}{c}{CIFAR-100-LT, IF = 0.01} \\
\cmidrule(lr){2-4} \cmidrule(lr){5-7} \cmidrule(lr){8-10}
& ACC (\%) & ARI (\%) & NMI (\%)
& ACC (\%) & ARI (\%) & NMI (\%)
& ACC (\%) & ARI (\%) & NMI (\%) \\
\midrule
Deep Clustering
& 24.3 (12.7) & 28.8 & 41.3
& 21.9 (18.9) & 12.2 & 35.2 
& 21.6 (15.3) & 12.9 & 35.7 \\

Semi-Supervised
& 52.6 (44.5) & 69.6 & 65.9
& 65.4 (65.0) & 48.4 & 67.1 
& 61.2 (69.0) & 45.3 & 63.3 \\

\bottomrule
\end{tabular}%
}
\end{table}

\begin{figure}[htbp]
    \centering
    \includegraphics[width=\linewidth]{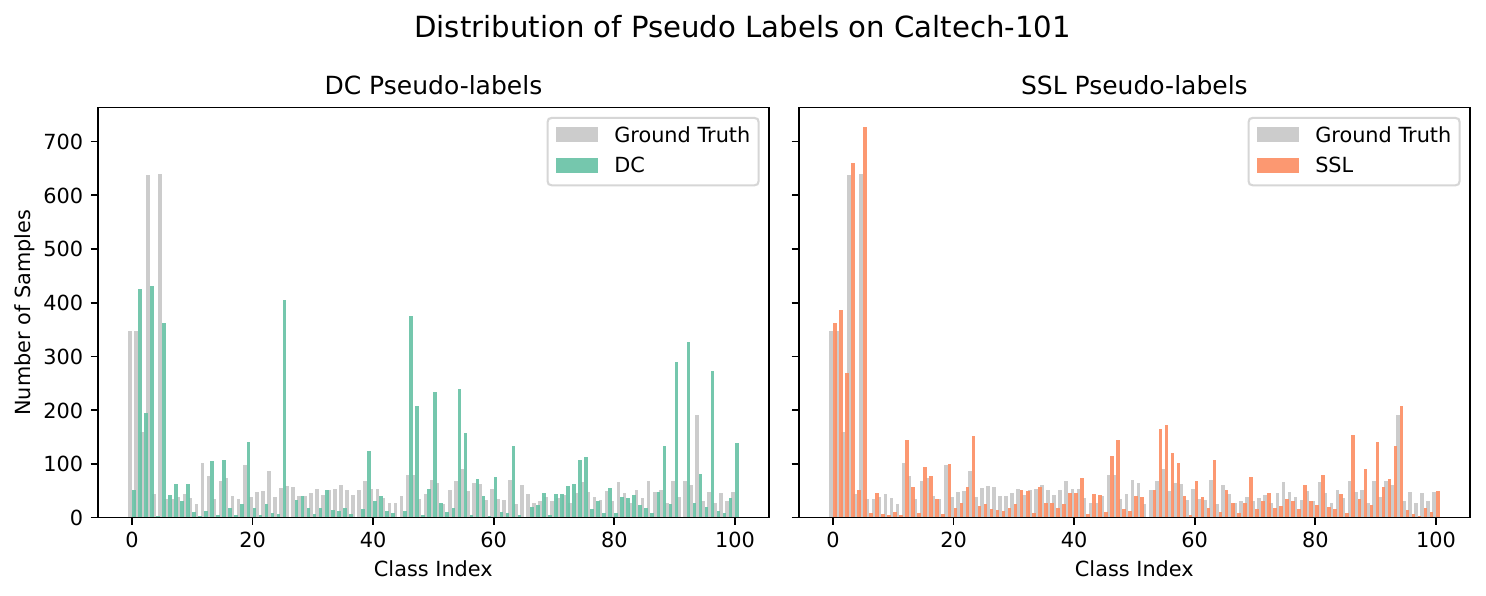}
    \caption{Class-wise distribution of assigned pseudo-labels on Caltech-101.}
    \label{fig:caltech_ps_dist}
\end{figure}

\begin{figure}[htbp]
    \centering
    \includegraphics[width=\linewidth]{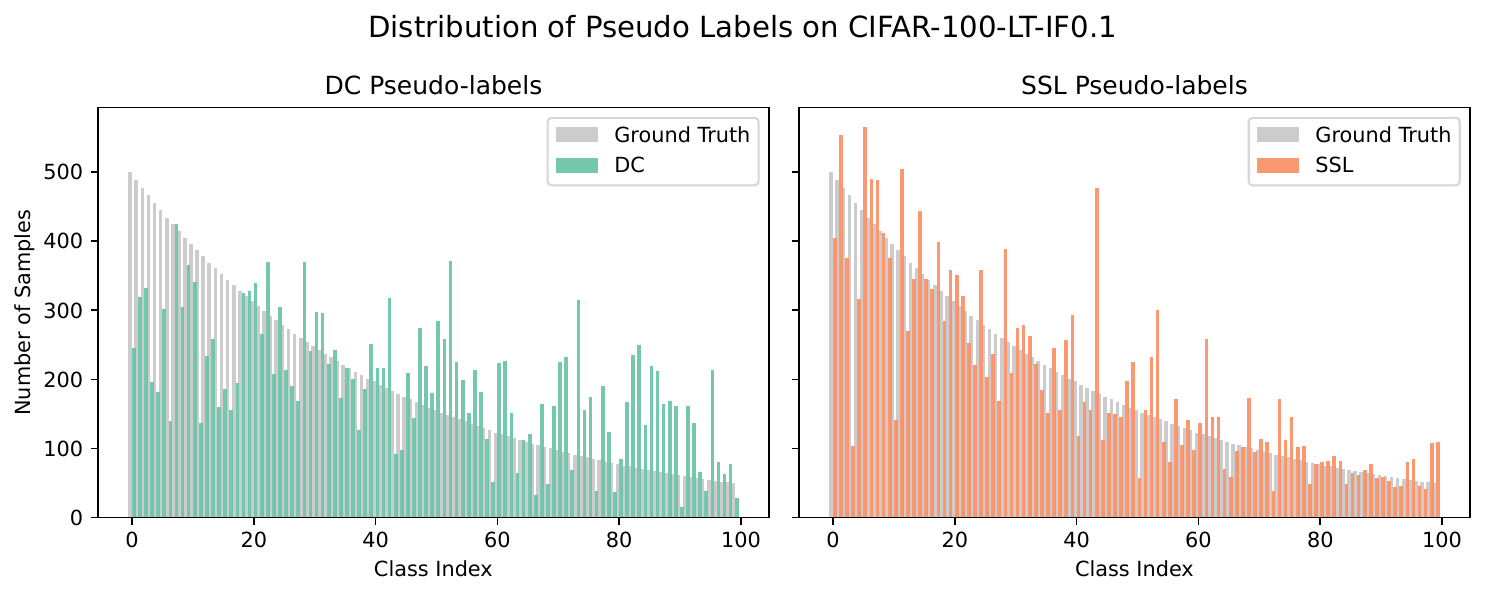}
    \caption{Class-wise distribution of assigned pseudo-labels on CIFAR-100-LT with an imbalance factor of 0.1.}
    \label{fig:cifar_lt_01_ps_dist}
\end{figure}

\begin{figure}[htbp]
    \centering
    \includegraphics[width=\linewidth]{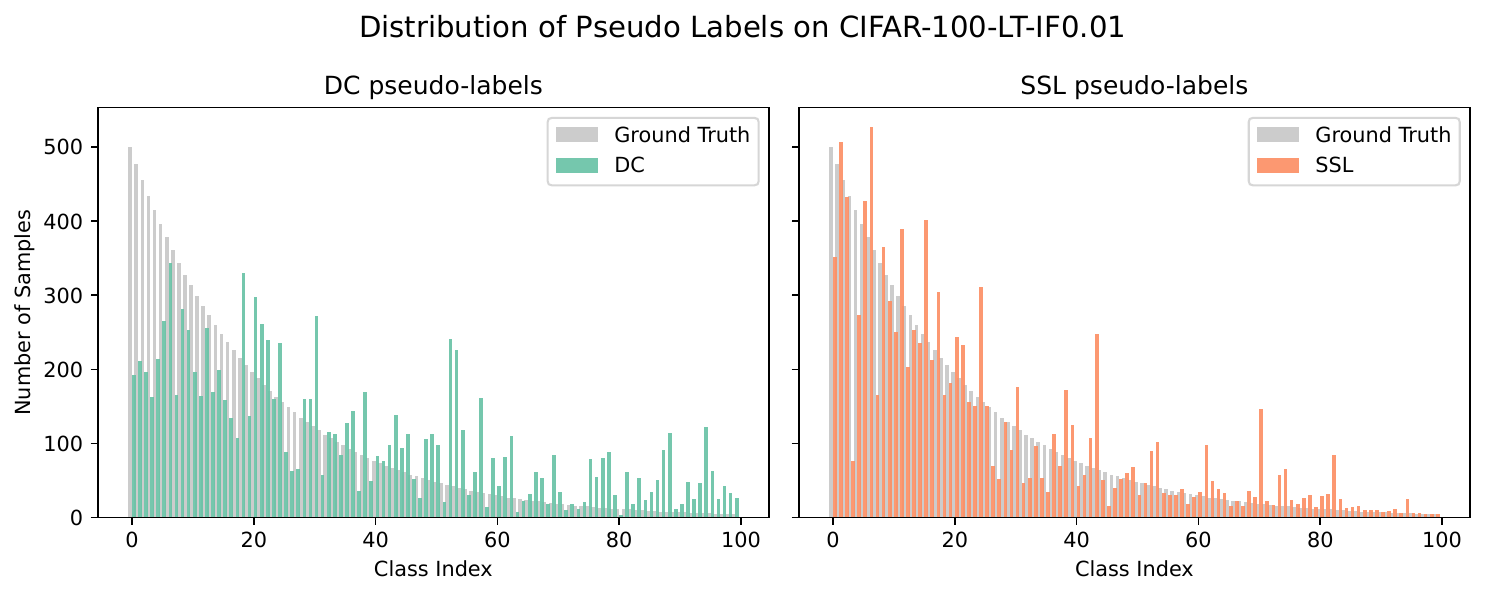}
    \caption{Class-wise distribution of assigned pseudo-labels on CIFAR-100-LT with an imbalance factor of 0.01.}
    \label{fig:cifar_lt_001_ps_dist}
\end{figure}

Figures~\ref{fig:caltech_ps_dist},~\ref{fig:cifar_lt_01_ps_dist}, and~\ref{fig:cifar_lt_001_ps_dist} show the corresponding class-wise pseudo-label distributions. The coreset results for the more severe CIFAR-100-LT setting, with an imbalance factor of 0.01, are reported separately in Appendix~\ref{appendix:severe_imbalance}.

\FloatBarrier

\subsection{How Pseudo-Label Errors Affect Pruning}
\label{appendix:pl_error_effects}

\subsubsection{Class-wise Error Structure}
\label{appendix:pl_classwise}

Mean pseudo-label accuracy can mask severe lower-tail errors. On SUN397, \ELFS{} and FixMatch reach mean pseudo-label accuracies of 38.1\% and 49.2\%, but worst-class accuracies of 0.00\% and 5.28\%. We trace this structure through scoring and selection on Food-101 at a 50\% pruning ratio.

\begin{table}[htbp]
\centering
\caption{Class-wise pseudo-label quality traced through scoring and selection. Food-101, 50\% pruning ratio.}
\label{tab:classwise_food}
\vspace{0.4em}
\resizebox{0.8\linewidth}{!}{%
\renewcommand{\arraystretch}{1.1}
\begin{tabular}{lcc}
\toprule
Metric & \ELFS{}-\AUM{} & \SemiAUM{} \\
\midrule
Mean pseudo-label accuracy & 42.81\% & 61.66\% \\
Worst-class pseudo-label accuracy & 0.40\% & 20.13\% \\
Class-level Spearman $\rho$ (PL accuracy vs.\ mean \AUM{} rank) & 0.893 & 0.968 \\
Overlap with GT-selected subset & 60.14\% & 68.26\% \\
Coreset test accuracy & 73.3\% & 73.9\% \\
\bottomrule
\end{tabular}%
}
\end{table}

Class-wise pseudo-label accuracy is strongly associated with mean \AUM{} rank, and correctly pseudo-labeled examples receive higher \AUM{} scores than incorrectly labeled ones even within the same ground-truth class. \SemiAUM{} also agrees more with ground-truth-based selection than \ELFS{}-\AUM{}: 68.26\% versus 60.14\% overlap, corresponding to Jaccard overlaps of 51.81\% and 43.00\%. The two selected subsets have similar ground-truth class-distribution total-variation distances (0.063 vs.\ 0.059), which suggests the difference arises from changes in score ordering and retained examples rather than from aggregate class balance.

On Food-101, these substantial differences in class-wise pseudo-label quality and selection agreement produce only a modest gap in overall accuracy (73.3\% vs.\ 73.9\%). To test whether the difference becomes more visible under class imbalance, we report lower-tail behavior on Caltech-101 at a 50\% pruning ratio.

\begin{table}[htbp]
\centering
\caption{Lower-tail pseudo-label and model accuracy under class imbalance. Caltech-101, 50\% pruning ratio.}
\label{tab:classwise_caltech}
\vspace{0.4em}
\resizebox{0.9\linewidth}{!}{%
\renewcommand{\arraystretch}{1.1}
\begin{tabular}{lcccc}
\toprule
Method & Worst-class PL acc. & Coreset test acc. & Worst-class model acc. & Bottom-5 class mean acc. \\
\midrule
\ELFS{}-\AUM{} & 0.00\% & 68.1\% & 0.00\% & 1.90\% \\
\SemiAUM{}     & 8.00\% & 70.6\% & 2.78\% & 5.52\% \\
\bottomrule
\end{tabular}%
}
\end{table}

On Caltech-101 the improvement in lower-tail pseudo-label quality is accompanied by better overall and lower-tail downstream performance, which suggests that the effect of pseudo-label-based scoring becomes more visible under class imbalance.

\FloatBarrier

\subsubsection{Controlled Pseudo-Label Noise}
\label{appendix:pl_noise}

The preceding analyses describe errors produced by actual pseudo-label generators. To isolate how label errors propagate into scoring and selection, we run a controlled experiment on CIFAR-100 at a 50\% pruning ratio. Starting from ground-truth labels, we randomly perturb labels to obtain target pseudo-label accuracies from 50\% to 90\%, and rerun the complete scoring, selection, and coreset-training pipeline at each level. Spearman correlation and subset overlap are measured against the corresponding fully supervised pruning run, which appears in the 100\% column.

\begin{table}[htbp]
\centering
\caption{Effect of controlled pseudo-label noise on score ordering, selected-subset membership, and downstream coreset accuracy. CIFAR-100, 50\% pruning ratio. Columns denote the pseudo-label accuracy of the perturbed labels; the 100\% column is the fully supervised reference run.}
\label{tab:pl_noise_sweep}
\vspace{0.4em}
\resizebox{0.9\linewidth}{!}{%
\renewcommand{\arraystretch}{1.1}
\begin{tabular}{llcccccc}
\toprule
Method & Metric & 50\% & 60\% & 70\% & 80\% & 90\% & 100\% (GT) \\
\midrule
\multirow{3}{*}{\SemiAUM}
 & Spearman correlation & 0.107 & 0.239 & 0.403 & 0.576 & 0.776 & 1.000 \\
 & Subset overlap       & 0.466 & 0.540 & 0.553 & 0.568 & 0.704 & 1.000 \\
 & Coreset accuracy     & 73.26 & 74.12 & 74.64 & 74.65 & 75.49 & 75.36 \\
\midrule
\multirow{3}{*}{\SemiDUAL}
 & Spearman correlation & $-0.083$ & $-0.090$ & 0.012 & 0.248 & 0.592 & 1.000 \\
 & Subset overlap       & 0.350 & 0.343 & 0.315 & 0.413 & 0.543 & 1.000 \\
 & Coreset accuracy     & 70.95 & 70.90 & 69.24 & 71.99 & 72.77 & 74.40 \\
\bottomrule
\end{tabular}%
}
\end{table}

Despite some non-monotonicity, particularly for \DUAL{} under severe noise, higher pseudo-label quality generally increases agreement with the ground-truth score ordering and the ground-truth selected subset, and also improves downstream coreset accuracy. Pseudo-label errors therefore do not act only through aggregate label accuracy: they reach the score ranking, the retained examples, and the final coreset performance.

Two points qualify this experiment. First, symmetric random corruption differs from the errors an SSL model actually makes: random corruption assigns incorrect labels independently of example difficulty, whereas SSL errors are structured by the model's representations and confidence. For the FixMatch pseudo-labels on CIFAR-100, the Spearman correlation between pseudo-label-based and ground-truth-based \DUAL{} scores is 0.655 overall, and 0.816 among correctly pseudo-labeled examples. The weak correlations observed under severe random corruption therefore do not transfer directly to the practical SSL setting.

Second, \SemiDUALBeta{} still selects a reasonable coreset even where the composite \DUAL{}-score correlation is near zero. This is because Beta sampling does not draw sampling probabilities from the composite score ranking; it assigns them from the prediction mean. The Spearman correlation between the pseudo-label-based and ground-truth-based prediction mean remains high, at 0.850 and 0.898 for 50\% and 60\% label accuracy respectively. The axis that Beta sampling actually uses thus retains ground-truth-consistent structure even when the composite score does not.

\FloatBarrier
\section{Experimental Details}
\label{appendix:exp_details}
We describe the compared methods, training protocols, and coreset-selection hyperparameters used in our experiments.

\subsection{Baselines}
\textbf{\Random} selects training examples uniformly at random. 
\textbf{\FullySup} uses ground-truth labels with \DUALBeta, where \DUAL{} estimates the example difficulty and uncertainty from early training stages, and Beta sampling adaptively selects a coreset according to the target pruning ratio.

\textbf{\Prototypicality} first clusters the data using $k$-means and measures each example's distance to its nearest cluster centroid; examples farther from centroids are treated as less prototypical and are prioritized for selection. 
\textbf{\ZCore} is another label-free baseline that leverages CLIP embeddings and iteratively selects a subset by balancing representativeness and redundancy. 
\textbf{\ELFSDINO} is a label-free selection baseline that applies the \ELFS{} pipeline using DINO features.
\textbf{\ELFSSelfEnc} is a variant of \ELFS{} that pretrains an encoder directly on the target dataset from scratch before performing deep clustering; this setting tests whether learning target-specific representations can improve label-free selection in challenging scenarios. We follow the original \ELFS{} and train the encoder with SwAV~\citep{caron2020unsupervised}. For CIFAR datasets, we use a CIFAR-adapted SwAV implementation as a reference~\citep{agarwalla_swav_cifar10}. For the remaining datasets, we use the ImageNet training hyperparameters reported in the original SwAV paper.

\textbf{\ScoreExtrap} assumes access to a small labeled subset, trains a supervised scorer on this subset, and extrapolates the resulting scores to the remaining unlabeled examples; we use the same initial label budget as our method for a fair comparison. 
We report two variants of our method. 
\textbf{\SemiAUMCutoff} uses SSL-generated pseudo-labels and applies \AUM-based double-end pruning with a cutoff strategy, following the selection principle of \ELFS. 
\textbf{\SemiDUALBeta} uses the same \DUALBeta{} selection rule as \FullySup.

\FloatBarrier

\subsection{Training Protocols}
We use a unified training protocol across datasets, while adjusting the model architecture and batch size according to dataset scale. 
We generate pseudo-labels using FixMatch, implemented with the publicly available USB codebase%
\footnote{\url{https://github.com/microsoft/Semi-supervised-learning.git}}.
The corresponding training hyperparameters are provided in Table~\ref{tab:fixmatch_hyperparams}.

For coreset selection, we first train a model on the full pseudo-labeled dataset, and then retrain the model from scratch on the selected coreset using the same base hyperparameters. 
For all datasets except ImageNet-1K, we halve the batch size at 80\% pruning and halve it again at 90\% pruning. 
We build our implementation on top of the publicly available \DUAL{} and \ELFS{} codebases%
\footnote{\url{https://github.com/behaapyy/dual-pruning}; \url{https://github.com/eltsai/elfs.git}}.
Unless otherwise specified, we follow their original training, scoring, and coreset selection procedures, and extend them to support pseudo-labeled datasets and our experimental settings.
The detailed training hyperparameters are summarized in Table~\ref{tab:training_hyperparams}.
For the main multi-run results, pseudo-labels are fixed while scoring and downstream training seeds vary. Additional full-pipeline seed sensitivity results on CIFAR-100 are reported in Appendix~\ref{app:seed_sensitivity}.

\begin{table}[htbp]
\centering
\caption{Training hyperparameters for pseudo-label generation. The same settings for CIFAR-10 and CIFAR-100 are also applied to their corrupted and long-tailed variants, i.e., CIFAR-100-C, CIFAR-100-LT.}
\label{tab:fixmatch_hyperparams}
\resizebox{0.6\linewidth}{!}{
\begin{tabular}{lcccc}
\toprule
Dataset & Architecture & Epochs & Batch Size & U Ratio \\
\midrule
CIFAR-10 & WRN-28-2 & 100 & 64 & 7 \\
CIFAR-100 & WRN-28-8 & 100 & 64 & 7 \\
Caltech-101 & ResNet-50 & 100 & 64 & 1\\
Food-101 & ResNet-50 & 50 & 64 & 1\\
SUN397 & ResNet-50 & 50 & 64 & 1\\
Tiny-ImageNet & ResNet-50 & 100 & 128 & 1\\
ImageNet-1K & ResNet-50 & 100 & 128 & 1\\
\bottomrule
\end{tabular}
}
\end{table}

\begin{table}[htbp]
\centering
\caption{Training hyperparameters used for coreset selection. The same settings for CIFAR-10 and CIFAR-100 are also applied to their corrupted and long-tailed variants, i.e., CIFAR-100-C, CIFAR-100-LT. We use CIFAR-style ResNet for CIFAR-10 and CIFAR-100, and use ImageNet-style ResNet for the others.}
\label{tab:training_hyperparams}
\renewcommand{\arraystretch}{1.2}
\resizebox{\linewidth}{!}{
\begin{tabular}{lcccccc}
\toprule
Dataset & Architecture & Epochs & Batch Size & LR & Optimizer & Weight Decay \\
\midrule
CIFAR-10 / CIFAR-100 
& ResNet18 & 200 & 128 & 0.1 & SGD, momentum 0.9 & $5 \times 10^{-4}$ \\

Food101 / SUN397 / Caltech101
& ResNet18 & 200 & 128 & 0.1 & SGD, momentum 0.9 & $1 \times 10^{-4}$ \\

ImageNet-1K / Tiny-ImageNet
& ResNet34 & 90 & 256 & 0.1 & SGD, momentum 0.9 & $1 \times 10^{-4}$ \\
\bottomrule
\end{tabular}
}
\end{table}

\FloatBarrier

\subsection{Selection and Search Hyperparameters}
Following \citet{zheng2025elfs}, we use a 90/10 train-validation split of the pseudo-labeled dataset for hyperparameter search, evaluating validation accuracy against pseudo-labels. This split is used only for hyperparameter search; the scorer is trained on the full pseudo-labeled pool. Once the hyperparameters are fixed, we select the final coreset from the entire training pool, including the 10\% validation subset. For \SemiAUMCutoff, we perform a grid search over the cutoff from 0 to $r$, where $r$ is the pruning ratio, with a step size of 0.1. For \SemiDUALBeta, we first fix $T$ based on the best performance at a 30\% pruning ratio, and then tune $c_D$ with the selected $T$. For \SemiDUALBeta{}, we additionally tune the uncertainty exponent $\gamma$, which reduces sensitivity to uncertainty estimates under noisy pseudo-labels. For CIFAR-10 and CIFAR-100, we search $\gamma \in \{0.5, 0.8, 1.0, 1.2\}$ and select $\gamma = 1.2$ based on validation performance; for ImageNet-1K, we evaluate $\gamma \in \{0.5, 1.0\}$ and select $\gamma = 0.5$. 

\begin{table}[htbp]
\centering
\caption{Cutoff ratios used for \SemiAUMCutoff. 
For each dataset and pruning ratio, we report the cutoff ratio used to remove low-AUM examples before selecting the final coreset.}
\vspace{0.4em}
\resizebox{0.7\textwidth}{!}{%
\begin{tabular}{llccccc}
\toprule
Category & Dataset & 30\% & 50\% & 70\% & 80\% & 90\% \\
\midrule

\multirow{2}{*}{Domain-specific}
& Food-101    & 0.1 & 0.3 & 0.5 & 0.6 & 0.8 \\
& SUN397     & 0.1 & 0.1 & 0.1 & 0.2 & 0.2 \\

\midrule
\multirow{2}{*}{Image-corrupted}
& CIFAR-100   & 0.1 & 0.2 & 0.4 & 0.5 & 0.7 \\
& Tiny-ImageNet & 0.2 & 0.4 & 0.6 & 0.7 & 0.8 \\

\midrule
\multirow{2}{*}{Long-tailed}
& CIFAR-100 (IF=0.1) & 0.1 & 0.4 & 0.5 & 0.6 & 0.8 \\
& Caltech-101  & 0.0 & 0.1 & 0.4 & 0.5 & 0.6 \\

\midrule
\multirow{3}{*}{Standard}
& CIFAR-10  & 0.0 & 0.0 & 0.1 & 0.2 & 0.4 \\
& CIFAR-100 & 0.1 & 0.2 & 0.4 & 0.6 & 0.7 \\
& ImageNet-1K & 0.1 & 0.3 & 0.5 & 0.5 & 0.7 \\

\bottomrule
\end{tabular}%
}
\end{table}

\vspace{10pt}
\begin{table}[htbp]
\centering
\caption{Hyperparameters used for \SemiDUALBeta. 
We report $(T, c_D)$, where $T$ is the score-computation epoch and $c_D$ is the Beta-sampling parameter reflecting dataset difficulty.}
\vspace{0.4em}
\resizebox{0.8\textwidth}{!}{%
\begin{tabular}{cc|cc|cc|ccc}
\toprule
\multicolumn{2}{c|}{Domain-specific} 
& \multicolumn{2}{c|}{Image-corrupted} 
& \multicolumn{2}{c|}{Long-tailed}
& \multicolumn{3}{c}{Standard} \\
\cmidrule(lr){1-2}
\cmidrule(lr){3-4}
\cmidrule(lr){5-6}
\cmidrule(lr){7-9}
\hspace{.25em}Food & SUN 
& C-100 & Tiny-IN
& C-100 & Caltech 
& C-10 & C-100 & IN-1K \\
\midrule
\hspace{.25em}$(30, 5)$ 
& $(30, 5)$ 
& $(70, 4)$ 
& $(50, 6)$ 
& $(30, 6)$ 
& $(30, 5)$ 
& $(30, 5.5)$ 
& $(30, 4)$ 
& $(60, 11)$ \\
\bottomrule
\end{tabular}%
}
\end{table}

\paragraph{Sensitivity to the uncertainty exponent.}
Table~\ref{tab:gamma_ablation} reports the effect of the selected uncertainty exponent on CIFAR-10 and CIFAR-100. The selected value changes average accuracy only modestly relative to $\gamma=1$, with the largest differences occurring at 90\% pruning.

\begin{table}[htbp]
\centering
\caption{Effect of the uncertainty exponent $\gamma$ on coreset accuracy. Parentheses give the change relative to $\gamma = 1$.}
\label{tab:gamma_ablation}
\vspace{0.4em}
\resizebox{0.8\linewidth}{!}{%
\renewcommand{\arraystretch}{1.15}
\begin{tabular}{llccccc}
\toprule
Dataset & $\gamma$ & 30\% & 50\% & 70\% & 80\% & 90\% \\
\midrule
\multirow{2}{*}{CIFAR-10}  & 1.0 & 95.4 & 95.1 & 92.6 & 90.8 & 85.8 \\
                           & 1.2 & 95.4 ($+$0.0) & 95.0 ($-$0.1) & 92.7 ($+$0.1) & 90.9 ($+$0.1) & 86.2 ($+$0.4) \\
\midrule
\multirow{2}{*}{CIFAR-100} & 1.0 & 77.2 & 73.8 & 67.2 & 62.7 & 52.3 \\
                           & 1.2 & 77.1 ($-$0.1) & 73.8 ($+$0.0) & 67.3 ($+$0.1) & 62.9 ($+$0.2) & 53.1 ($+$0.8) \\
\bottomrule
\end{tabular}%
}
\end{table}

\FloatBarrier
\section{Additional Evaluations and Comparisons}
\label{appendix:additional_evaluation}
We provide complementary evaluations under more severe class imbalance and against dynamic pruning methods.

\FloatBarrier

\subsection{Severe Class Imbalance}
\label{appendix:severe_imbalance}
We extend the long-tailed evaluation to CIFAR-100-LT with an imbalance factor of 0.01. The balanced test set is unchanged, and the full training set achieves 41.7\% test accuracy. Pseudo-label quality and class distributions for this setting are analyzed in Appendix~\ref{appendix:dc_ssl_embeddings_longtailed}.

\begin{table}[htbp]
\centering
\caption{Coreset performance on long-tailed CIFAR-100 with imbalance factor 0.01. We use the original CIFAR-100 test set, which is balanced. Results are averaged over five runs. The test accuracy for the full train dataset is 41.7\%.}
\vspace{0.2em}
\renewcommand{\arraystretch}{1.0}
\label{tab:cifar100_lt_001}
\resizebox{0.75\textwidth}{!}{%
\begin{tabular}{llccccc}
\toprule
Label Usage & Method & 30\% & 50\% & 70\% & 80\% & 90\% \\
\midrule
With Labels & \FullySup{}
& 37.0 & 32.3 & 21.9 & 19.4 & 15.4 \\
\midrule
\multirow{5}{*}{Without Labels}
& \Random{}
& 22.9 & 11.5 & 5.9 & 4.0 & 2.5 \\
& \Prototypicality{}
& 34.9 & 29.2 & 19.8 & 14.7 & 11.0  \\
& \ELFSDINO{}
& \uline{35.0} & \textbf{31.8} & \uline{24.0} & \uline{19.9} & 12.4 \\
& \ELFSSelfEnc{}
& 34.8 & 29.3 & 22.6 & 16.8 & 12.4 \\
& \ZCore{}
& 32.5 & 28.8 & 20.0 & 16.1 & 12.6 \\
\midrule
\multirow{3}{*}{With 10\% Labels}
& \ScoreExtrap{}
& 34.7 & 28.4 & 20.0 & 16.1 & 12.1 \\
& \hl\SemiAUMCutoff{}
& \hl33.8 & \hl30.6 & \hl\textbf{24.7} & \hl\textbf{20.1} & \hl\textbf{14.5} \\
& \hl\SemiDUALBeta{}
& \hl\textbf{38.0} & \hl\uline{31.0} & \hl22.4 & \hl18.7 & \hl\uline{14.4} \\
\bottomrule
\end{tabular}%
} 
\end{table}

Table~\ref{tab:cifar100_lt_001} shows that the relative performance of the two scoring criteria varies with the pruning ratio. \SemiDUALBeta{} achieves the highest accuracy among the label-free and label-efficient methods at 30\% pruning, whereas \SemiAUMCutoff{} performs best at 70--90\% pruning. \ELFSDINO{} remains strongest at 50\% pruning. These results further illustrate that the preferred scoring criterion depends on the data regime and pruning ratio.

\FloatBarrier

\subsection{Comparison with Dynamic Pruning}
\label{appendix:dynamic_pruning}

\subsubsection{Comparison Setting}
\label{appendix:dynamic_pruning_setting}

Dynamic and online pruning methods such as InfoBatch~\citep{qin2024infobatch} and Partial Forward Blocking (PFB)~\citep{wu2025partial} also reduce the effective amount of data used for training, but they operate under a different data-access and supervision setting from ours. They prune online while retaining access to the full pool throughout training, so an example discarded at one stage can be reintroduced later. In this comparison, the dynamic baselines train on pseudo-labels from the initial pseudo-label pool. \SemiPrune{} uses this pool for scoring and selection, then trains on ground-truth labels for the selected coreset without revisiting the discarded data.

We match the number of optimizer updates and the number of retained examples per update. \SemiPrune{} trains its retained subset for 200 epochs. PFB enlarges its candidate batch so that, after online pruning, its selected batch size and its number of updates per epoch match the static setting. All methods consequently process the same total number of retained training examples. The downstream supervision and data-access settings differ, however: the dynamic baselines use pseudo-labels and retain full-pool access, whereas \SemiPrune{} uses ground-truth labels and a fixed coreset. The comparison therefore reflects the combined effects of selection, supervision, and data access.

\subsubsection{Coreset Performance}
\label{appendix:dynamic_pruning_performance}
Table~\ref{tab:dynamic_pruning} compares static and dynamic pruning under the matched update budget described above.

\begin{table}[htbp]
\centering
\caption{Static versus dynamic pruning under matched optimizer updates and matched retained examples per update. Dynamic baselines train on pseudo-labels; \SemiPrune{} trains on ground-truth labels for the selected coreset. The first row gives the PFB candidate/selected batch sizes used at each pruning ratio. Best result per column in bold.}
\label{tab:dynamic_pruning}
\vspace{0.4em}
\resizebox{0.95\linewidth}{!}{%
\renewcommand{\arraystretch}{1.15}
\begin{tabular}{llcccc}
\toprule
Dataset & Method & 30\% & 50\% & 70\% & 90\% \\
\midrule
PFB config & Candidate / selected batch & 183 / 128 & 256 / 128 & 427 / 128 & 320 / 32 \\
\midrule
\multirow{4}{*}{CIFAR-100}
 & InfoBatch   & 72.78 $\pm$ 0.17 & 72.09 $\pm$ 0.46 & 70.70 $\pm$ 0.28 & 52.18 $\pm$ 2.35 \\
 & PFB         & 72.62 $\pm$ 0.20 & 72.45 $\pm$ 0.25 & \textbf{71.43 $\pm$ 0.07} & \textbf{66.87 $\pm$ 0.83} \\
 & \SemiAUM{}  & \textbf{77.40 $\pm$ 0.23} & 73.50 $\pm$ 0.10 & 67.70 $\pm$ 0.26 & 53.90 $\pm$ 0.23 \\
 & \SemiDUAL{} & 77.10 $\pm$ 0.18 & \textbf{73.80 $\pm$ 0.27} & 67.30 $\pm$ 0.24 & 53.10 $\pm$ 0.49 \\
\midrule
\multirow{4}{*}{CIFAR-100-C}
 & InfoBatch   & 58.29 $\pm$ 0.80 & 56.66 $\pm$ 0.23 & 53.80 $\pm$ 0.83 & 31.56 $\pm$ 2.84 \\
 & PFB         & 58.23 $\pm$ 0.22 & 57.57 $\pm$ 0.14 & 55.78 $\pm$ 0.33 & \textbf{55.30 $\pm$ 1.30} \\
 & \SemiAUM{}  & 73.10 $\pm$ 0.19 & 70.00 $\pm$ 0.31 & \textbf{63.90 $\pm$ 0.59} & 52.00 $\pm$ 0.31 \\
 & \SemiDUAL{} & \textbf{73.30 $\pm$ 0.28} & \textbf{70.40 $\pm$ 0.39} & 63.80 $\pm$ 0.76 & 47.50 $\pm$ 0.81 \\
\bottomrule
\end{tabular}%
}
\end{table}

The results show a trade-off between the static and dynamic regimes rather than a uniform ordering. At moderate pruning ratios (30--50\%) the retained subset is still large enough to preserve example coverage, and \SemiPrune{} benefits from acquiring ground-truth labels for the retained examples, outperforming both dynamic methods by a wide margin on CIFAR-100-C. Under aggressive pruning the picture reverses: repeatedly training on a single fixed subset limits example diversity, whereas PFB can distribute the same update budget over different examples drawn from the full pool. PFB is accordingly stronger at 90\% pruning on both datasets. These results illustrate the trade-off between ground-truth supervision on a fixed coreset and pseudo-label supervision with continued access to the full pool.

\FloatBarrier

\FloatBarrier
\section{Additional Ablations}
\label{appendix:ablations}
We extend the initial-label-budget analysis to additional pruning
ratios and examine the effects of the SSL algorithm, pseudo-label
generator architecture, and scoring criterion.

\subsection{Effect of the Initial Label Budget}
\label{appendix:label_budget}
\begin{figure}[htbp]
    \centering
    \includegraphics[width=0.9\linewidth]{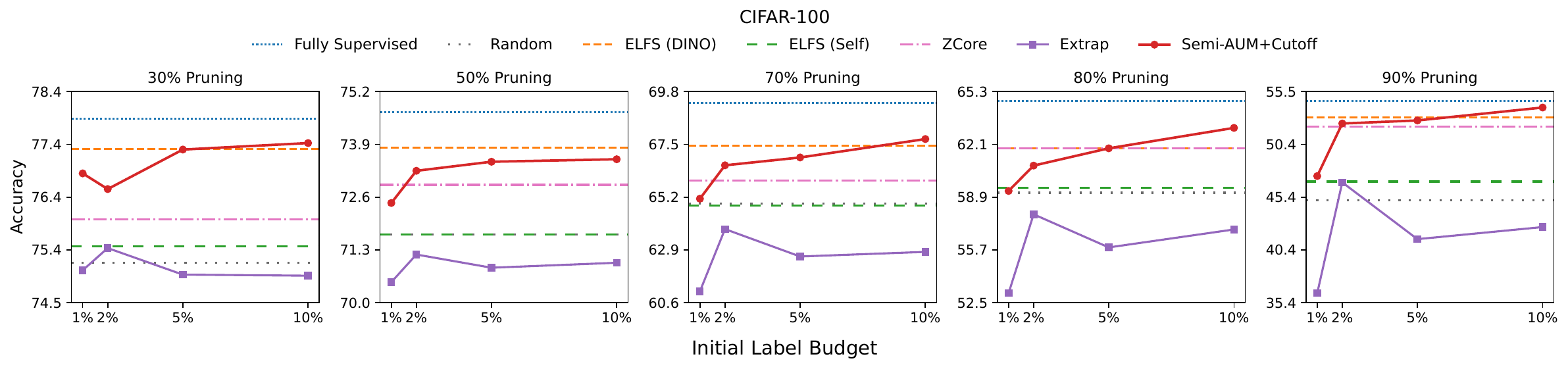}

    \vspace{0.5em}

    \includegraphics[width=0.9\linewidth]{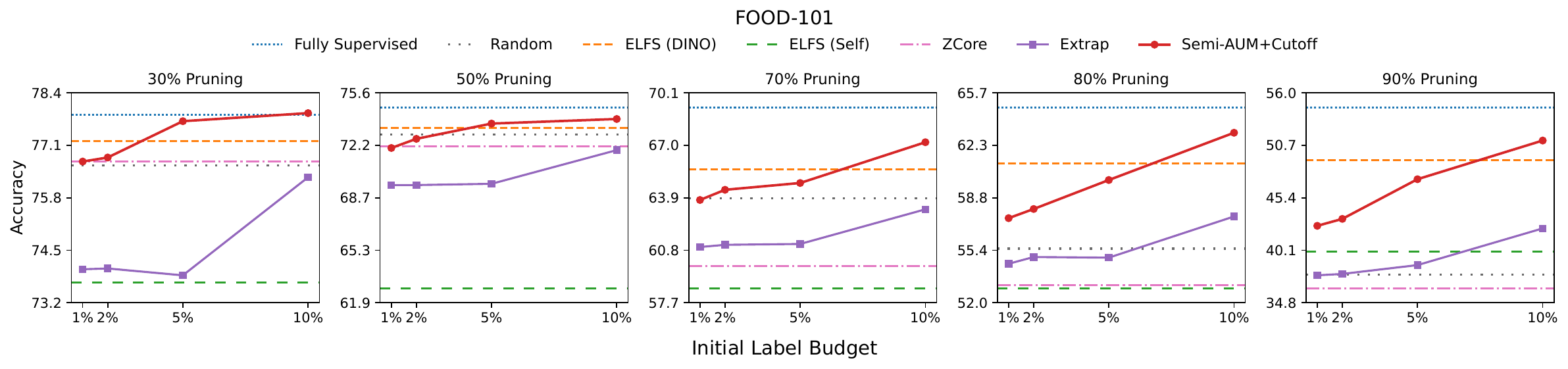}
    \caption{Effect of the initial label budget on coreset performance
    on CIFAR-100 (top) and Food-101 (bottom).}
    \label{fig:label_budget_ablation}
\end{figure}

We extend the initial-label-budget analysis in the main text to all evaluated pruning ratios on CIFAR-100 and Food-101 (Figure~\ref{fig:label_budget_ablation}). Our \SemiAUM{} consistently performs better than \ScoreExtrap{} under the same initial annotation budget. With 5\% annotated samples, \SemiAUM{} can outperform the strongest label-free baseline, \ELFS. 

\FloatBarrier

\subsection{Generality Across Design Choices}
\label{appendix:pl_generator_robustness}

\subsubsection{SSL Algorithm}
\label{appendix:ssl_generality}

To test whether the results depend on the specific SSL algorithm, we replace FixMatch with FlexMatch on CIFAR-100 while holding the scoring and downstream protocols fixed.

\begin{table}[htbp]
\centering
\caption{Effect of the SSL method used for pseudo-label generation. CIFAR-100, with scoring and downstream protocols held fixed.}
\label{tab:ssl_generality}
\vspace{0.4em}
\resizebox{0.9\linewidth}{!}{%
\renewcommand{\arraystretch}{1.15}
\begin{tabular}{llcccccc}
\toprule
Score & SSL method & PL acc. & 30\% & 50\% & 70\% & 80\% & 90\% \\
\midrule
\SemiAUM{}  & FixMatch  & 74.1 & 77.4 $\pm$ 0.23 & 73.5 $\pm$ 0.10 & 67.7 $\pm$ 0.26 & 63.1 $\pm$ 0.21 & 53.9 $\pm$ 0.23 \\
\SemiAUM{}  & FlexMatch & 77.8 & 77.3 $\pm$ 0.23 & 74.0 $\pm$ 0.11 & 68.1 $\pm$ 0.24 & 63.2 $\pm$ 0.37 & 54.2 $\pm$ 0.69 \\
\midrule
\SemiDUAL{} & FixMatch  & 74.1 & 77.1 $\pm$ 0.18 & 73.8 $\pm$ 0.27 & 67.3 $\pm$ 0.24 & 62.9 $\pm$ 0.31 & 53.1 $\pm$ 0.49 \\
\SemiDUAL{} & FlexMatch & 77.8 & 77.2 $\pm$ 0.35 & 73.8 $\pm$ 0.30 & 67.3 $\pm$ 0.29 & 62.8 $\pm$ 0.37 & 52.7 $\pm$ 0.42 \\
\bottomrule
\end{tabular}%
}
\end{table}

FlexMatch raises pseudo-label accuracy from 74.1\% to 77.8\%. With FlexMatch, \SemiAUM{} is comparable to or better than its FixMatch counterpart and continues to match or outperform \ELFS{} at every pruning ratio. \SemiDUAL{} stays comparable to both its FixMatch counterpart and \ELFS{}, but does not improve consistently. Higher aggregate pseudo-label accuracy therefore does not translate into uniform gains for every scoring criterion, while the framework's effectiveness is not specific to FixMatch.

\FloatBarrier

\subsubsection{Pseudo-Label Generator Architecture}
\label{appendix:architecture}

The pseudo-label generator and the scoring model play different roles. The SSL model is used only to produce proxy labels; its features, confidence values, and training dynamics are never consumed by \SemiAUM{} or \SemiDUAL{}. Following common SSL benchmark practice we use WRN-28-8 for CIFAR-100 and ResNet-50 for the larger datasets, rather than matching the generator to the downstream scoring model, which is trained from scratch on the proxy-labeled pool. Architectural matching is therefore not a requirement of the framework. We nevertheless evaluate a matched setting by replacing the WRN-28-8 FixMatch model with ResNet-18, the architecture also used for scoring and final subset training, holding the rest of the protocol fixed.

\begin{table}[htbp]
\centering
\caption{Matching the pseudo-label generator architecture to the scoring and subset-training model. CIFAR-100.}
\label{tab:architecture_matching}
\vspace{0.4em}
\resizebox{0.9\linewidth}{!}{%
\renewcommand{\arraystretch}{1.15}
\begin{tabular}{llcccccc}
\toprule
PL generator & Method & PL acc. & 30\% & 50\% & 70\% & 80\% & 90\% \\
\midrule
WRN-28-8  & \SemiAUM{}  & 74.1 & 77.4 $\pm$ 0.23 & 73.5 $\pm$ 0.10 & 67.7 $\pm$ 0.26 & 63.1 $\pm$ 0.21 & 53.9 $\pm$ 0.23 \\
ResNet-18 & \SemiAUM{}  & 68.4 & 77.4 $\pm$ 0.05 & 73.4 $\pm$ 0.64 & 66.9 $\pm$ 0.78 & 61.0 $\pm$ 0.50 & 52.7 $\pm$ 2.25 \\
\midrule
WRN-28-8  & \SemiDUAL{} & 74.1 & 77.1 $\pm$ 0.18 & 73.8 $\pm$ 0.27 & 67.3 $\pm$ 0.24 & 62.9 $\pm$ 0.31 & 53.1 $\pm$ 0.49 \\
ResNet-18 & \SemiDUAL{} & 68.4 & 77.3 $\pm$ 0.29 & 73.6 $\pm$ 0.42 & 66.2 $\pm$ 1.03 & 61.4 $\pm$ 0.82 & 51.5 $\pm$ 0.84 \\
\bottomrule
\end{tabular}%
}
\end{table}

Matching does not give a consistent improvement here. The ResNet-18 generator lowers pseudo-label accuracy from 74.1\% to 68.4\%; coreset performance stays comparable at 30--50\% pruning but declines more noticeably at higher ratios. This does not establish that architecture matching is generally irrelevant, but it shows no systematic benefit when matching reduces proxy-label quality.

\FloatBarrier
\subsubsection{Scoring Criterion}
\label{appendix:scoring_criterion}
\label{sec:scoring_metric_generality}
We finally examine whether the benefit of SSL-based pseudo-labeling
extends beyond the scoring criteria used in our main variants.
Specifically, we compare SSL-based and deep-clustering-based
pseudo-labels under Forgetting, EL2N, and AUM.

\begin{table}[htbp]
\centering
\caption{Pseudo-labeling effectiveness across scoring metrics.
SSL-based pseudo-labels yield consistent gains on Food-101 and remain
competitive on CIFAR-100.}
\label{tab:score_ablations}
\resizebox{0.85\linewidth}{!}{%
\renewcommand{\arraystretch}{1.15}
\begin{tabular}{llccccc|ccccc}
\toprule
\multirow{2}{*}{Score}
& \multirow{2}{*}{Pseudo-labels}
& \multicolumn{5}{c}{Food-101}
& \multicolumn{5}{c}{CIFAR-100} \\
\cmidrule(lr){3-7} \cmidrule(lr){8-12}
& & 30\% & 50\% & 70\% & 80\% & 90\%
& 30\% & 50\% & 70\% & 80\% & 90\% \\
\midrule

\multirow{2}{*}{Forgetting}
& Deep Clust.
& 77.4 & 73.0 & 64.8 & 60.7 & 48.7
& 77.1 & \textbf{74.2} & \textbf{67.5} & 62.0 & 51.8 \\
& Semi-sup.
& \textbf{77.7} & \textbf{73.4} & \textbf{66.4} & \textbf{62.0} & \textbf{52.5}
& \textbf{77.3} & 73.6 & 66.9 & \textbf{62.6} & \textbf{53.3} \\
\midrule

\multirow{2}{*}{EL2N}
& Deep Clust.
& 77.2 & 72.9 & 65.2 & 59.8 & 48.6
& 76.7 & 72.9 & 65.9 & 61.0 & 51.1 \\
& Semi-sup.
& \textbf{77.5} & \textbf{73.0} & \textbf{65.5} & \textbf{60.4} & \textbf{50.9}
& \textbf{77.5} & \textbf{73.2} & \textbf{66.8} & \textbf{61.4} & \textbf{51.7} \\
\midrule

\multirow{2}{*}{AUM}
& Deep Clust.
& 77.2 & 73.3 & 65.6 & 61.1 & 49.2
& 77.3 & \textbf{73.8} & 67.4 & 61.9 & 53.1 \\
& Semi-sup.
& \textbf{77.9} & \textbf{73.9} & \textbf{67.2} & \textbf{63.1} & \textbf{51.2}
& \textbf{77.4} & 73.5 & \textbf{67.7} & \textbf{63.1} & \textbf{53.9} \\
\bottomrule
\end{tabular}%
}
\end{table}

As shown in Table~\ref{tab:score_ablations}, SSL-based pseudo-labeling
improves coreset accuracy over deep clustering across all evaluated
scoring metrics and pruning ratios on Food-101.
On CIFAR-100, the two approaches perform more comparably, with
SSL-based pseudo-labeling achieving similar or better accuracy in most settings.
The larger gains on Food-101 are consistent with the greater value of
target-class supervision when pretrained representations are less
aligned with the target data.
These results show that the benefit of SSL-based pseudo-labeling
generalizes across different training-dynamics-based scoring criteria.

\FloatBarrier
\subsection{Full-Pipeline Seed Sensitivity}
\label{app:seed_sensitivity}

On CIFAR-100 with a 10\% initial label budget, we additionally repeat the full pipeline with five independent seeds. Each run independently resamples the initial labeled subset and repeats FixMatch training, scoring, coreset selection, and downstream training. Table~\ref{tab:full_pipeline_seeds} reports the resulting coreset test accuracy.

\begin{table}[htbp]
\centering
\caption{Full-pipeline seed sensitivity on CIFAR-100. Values are coreset test accuracy (\%), reported as mean $\pm$ sample standard deviation over five independent full-pipeline runs. Columns indicate pruning ratios.}
\label{tab:full_pipeline_seeds}
\vspace{0.4em}
\resizebox{0.9\linewidth}{!}{%
\renewcommand{\arraystretch}{1.15}
\begin{tabular}{lccccc}
\toprule
Method & 30\% & 50\% & 70\% & 80\% & 90\% \\
\midrule
\SemiAUM{} & 77.4 $\pm$ 0.15 & 73.6 $\pm$ 0.42 & 67.4 $\pm$ 0.54 & 62.3 $\pm$ 0.65 & 53.2 $\pm$ 0.50 \\
\SemiDUAL{} & 77.2 $\pm$ 0.28 & 73.7 $\pm$ 0.30 & 67.0 $\pm$ 0.26 & 62.4 $\pm$ 0.56 & 52.9 $\pm$ 0.22 \\
\bottomrule
\end{tabular}%
}
\end{table}

Across these five runs, pseudo-label accuracy is $74.04 \pm 0.13\%$, and downstream accuracy standard deviations are at most 0.65 percentage points. This additional analysis is limited to CIFAR-100; the main results use fixed pseudo-labels and vary scoring and downstream training seeds.
\FloatBarrier

\FloatBarrier
\section{Efficiency and Practical Considerations}
\label{appendix:computation_cost}
We report end-to-end computational costs, examine the pseudo-label generation stage across datasets, and discuss the practical trade-offs between the two scoring variants.

\subsection{End-to-End Cost}
\label{appendix:end_to_end_cost}

We first report the complete point-of-use cost, including the SSL stage required by \SemiPrune{}. The comparison covers proxy-label acquisition, dynamics collection, selection search, and final subset training. A downloaded DINO encoder has zero encoder-training cost at the point of use, so we separate \ELFSDINO{} from \ELFSSelfEnc{}, which trains SwAV on the target data. All timings were measured on NVIDIA RTX A6000 GPUs and are reported as elapsed wall-clock time. For Food-101, FixMatch proxy acquisition used four GPUs. The totals sum the elapsed times of the pipeline stages.

\begin{table}[htbp]
\centering
\caption{Stage-by-stage point-of-use cost on Food-101 at a 50\% pruning ratio. $\dagger$~The downloaded DINO encoder's pretraining cost is excluded. $\ddagger$~Search over six cutoff ratios, $\{0, 0.1, 0.2, 0.3, 0.4, 0.5\}$, including candidate-subset retraining. $\S$~Search over four $c_D$ candidates, $\{3, 4, 5, 6\}$, including candidate-subset retraining.}
\label{tab:end_to_end_food}
\vspace{0.4em}
\resizebox{\linewidth}{!}{%
\renewcommand{\arraystretch}{1.15}
\begin{tabular}{lccccccc}
\toprule
Method & \makecell{Initial\\labels} & \makecell{Proxy\\acquisition} & \makecell{Dynamics\\collection} & \makecell{Score/\\search} & \makecell{Subset\\training} & Total & Acc. \\
\midrule
\ELFSDINO{}    & 0\%   & 1h 36m$^\dagger$ & \makecell{4h 29m\\(200 ep)} & 19h 36m$^\ddagger$ & 3h 16m & \textbf{28h 57m} & 73.3\% \\
\ELFSSelfEnc{} & 0\%   & 13h 17m          & \makecell{4h 29m\\(200 ep)} & 19h 36m$^\ddagger$ & 3h 16m & \textbf{40h 38m} & 62.8\% \\
\ZCore{}       & 0\%   & $\sim$4m         & --              & $\sim$1m            & 3h 16m & \textbf{3h 21m}  & 72.1\% \\
\ScoreExtrap{} & 10\%  & --               & \makecell{$\sim$55m\\(200 ep)} & $\sim$15m       & 3h 16m & \textbf{4h 26m}  & 71.8\% \\
\SemiAUM{}     & 10\%  & 13h 42m          & \makecell{4h 47m\\(200 ep)} & 19h 36m$^\ddagger$ & 3h 16m & \textbf{41h 21m} & 73.9\% \\
\SemiDUAL{}    & 10\%  & 13h 42m          & \makecell{$\sim$43m\\(30 ep)} & 13h 04m$^\S$ & 3h 16m & \textbf{30h 45m} & 73.4\% \\
\bottomrule
\end{tabular}%
}
\end{table}

The score/search stage uses \AUM{} with double-end pruning for \ELFS{} and \SemiAUM{}, and \DUAL{} with Beta sampling for \SemiDUAL{}. Dynamics collection trains a separate scoring model on the proxy-labeled pool; all methods use the same downstream architecture and the same final subset-training schedule. Compared with \ELFSDINO{}, \SemiAUM{} and \SemiDUAL{} require a 10\% initial label budget and respectively add 12h 24m and 1h 48m of point-of-use time. We therefore do not claim universal compute savings.

Food-101 is a setting where the SSL stage is itself relatively expensive. For completeness we also report the end-to-end comparison on clean CIFAR-100, where the target distribution is well matched to the pretraining distribution.

\begin{table}[htbp]
\centering
\caption{Stage-by-stage point-of-use cost on clean CIFAR-100 at a 50\% pruning ratio. Footnote markers follow Table~\ref{tab:end_to_end_food}, except that $\S$ here denotes a search over four $c_D \in \{3, 4, 5, 6\}$ followed by $\gamma \in \{0.5, 0.8, 1.0, 1.2\}$, including candidate-subset retraining.}
\label{tab:end_to_end_cifar100}
\vspace{0.4em}
\resizebox{\linewidth}{!}{%
\renewcommand{\arraystretch}{1.15}
\begin{tabular}{lccccccc}
\toprule
Method & \makecell{Initial\\labels} & \makecell{Proxy\\acquisition} & \makecell{Dynamics\\collection} & \makecell{Score/\\search} & \makecell{Subset\\training} & Total & Acc. \\
\midrule
\ELFS{}     & 0\%  & 30m$^\dagger$ & \makecell{45m\\(200 ep)} & $\sim$2h 15m$^\ddagger$ & $\sim$20m & \textbf{$\sim$3h 50m} & 73.8\% \\
\SemiAUM{}  & 10\% & 4h            & \makecell{45m\\(200 ep)} & $\sim$2h 15m$^\ddagger$ & $\sim$20m & \textbf{$\sim$7h 20m} & 73.5\% \\
\SemiDUAL{} & 10\% & 4h            & \makecell{8m\\(30 ep)}   & $\sim$3h 10m$^\S$   & $\sim$20m & \textbf{$\sim$7h 38m} & 73.8\% \\
\bottomrule
\end{tabular}%
}
\end{table}

The additional target-specific SSL cost is most justified when pretrained representations are poorly aligned with the target distribution. On a clean benchmark such as CIFAR-100 the accuracy gain relative to the additional cost is limited, which is consistent with the positioning of \SemiPrune{} as an annotation--accuracy--compute trade-off rather than a compute reduction.

\FloatBarrier

\subsection{Cost of Pseudo-Label Generation}
\label{appendix:computation_time}
To examine the proxy-acquisition stage in more detail, we compare methods that require training on the target dataset for pseudo-label generation. All experiments were conducted on NVIDIA RTX A6000 GPUs.
\ELFS{} also considers a self-trained SwAV encoder, which requires target-dataset self-supervised training before clustering. 
In contrast, our method trains an SSL model directly on the target dataset using a small labeled subset and the remaining unlabeled examples.

\begin{table}[htbp]
\centering
\caption{Computation time for target-dataset pseudo-label generation. 
We compare target-dataset self-supervised encoder training used by SwAV with SSL-based pseudo-labeling methods. 
We report representative GPU-heavy benchmarks and omit smaller auxiliary settings from this comparison.}
\label{tab:pseudo_label_generation_time}
\vspace{0.4em}
\resizebox{0.8\linewidth}{!}{%
\renewcommand{\arraystretch}{1.15}
\begin{tabular}{lllcccc}
\toprule
Method & Dataset & Backbone & Epochs & GPUs & Wall-clock Time & GPU-hours \\
\midrule
SwAV & CIFAR-100 & ResNet-50 & 500 & 1  & 11.8 h  & 11.80 \\
SwAV & Food-101 & ResNet-50 & 200 & 4  & 12.06 h & 48.24 \\
SwAV & SUN397 & ResNet-50 & 200 & 8  & 9.57 h  & 76.56 \\
SwAV & Tiny-ImageNet & ResNet-50 & 200 & 4 & 30.40 h & 121.59 \\
SwAV & ImageNet-1K & ResNet-50 & 800 & 64 & 39.84 h & 2549.76 \\
\midrule
FixMatch & CIFAR-100 & WRN-28-8 & 100 & 1 & 4.00 h & 4.00 \\
FixMatch & Food-101 & ResNet-50 & 50 & 4 & 13.65 h & 54.60 \\
FixMatch & SUN397 & ResNet-50 & 50  & 4 & 13.82 h & 55.28 \\
FixMatch & Tiny-ImageNet & ResNet-50 & 100 & 4 & 7.3 h & 29.28 \\
FixMatch & ImageNet-1K & ResNet-50 & 100 & 8 & 32.88 h & 263.04 \\
\bottomrule
\end{tabular}
}
\end{table}

\subsection{Choosing Between \SemiAUMCutoff{} and \SemiDUALBeta{}}
\label{appendix:scorer_choice}

\SemiAUMCutoff{} and \SemiDUALBeta{} are reported as separate instantiations of the framework, not as a best-of-two result selected after observing test performance. We do not claim a universally validated \emph{a priori} rule for choosing between them, but the observed pattern admits a consistent reading based on what a low score represents in each setting.

Under image corruption, severely degraded examples tend to keep low or negative margins over many epochs, so a margin-based criterion with a cutoff is well suited to identifying and removing persistently unreliable examples. In long-tailed data a low margin may instead belong to a valid tail-class example that is learned late because it is underrepresented, and a global \AUM{} cutoff can then discard rare but informative examples. \SemiDUALBeta{} captures variation in the target probability over training while emphasizing low-confidence examples, and Beta sampling supplements those examples with moderate-difficulty ones. This may help retain informative tail examples, but the relative advantage depends on the imbalance level and pruning ratio. For example, on CIFAR-100-LT with an imbalance factor of 0.01 (Table~\ref{tab:cifar100_lt_001}), \SemiDUALBeta{} performs best among the label-free and label-efficient methods at 30\% pruning, whereas \SemiAUMCutoff{} performs best at 70--90\%. Thus, the reported long-tail results support a conditional empirical trend rather than a general preference for \SemiDUALBeta{} across long-tailed datasets.

Deployment cost provides a second practical consideration. \SemiDUALBeta{} requires less computation for dynamics collection in the reported comparisons, which can be useful when scoring is repeated. The end-to-end benefit also depends on pseudo-label generation and hyperparameter search: Appendix~\ref{appendix:end_to_end_cost} shows lower total cost for \SemiDUALBeta{} than for \SemiAUMCutoff{} on Food-101, but not on CIFAR-100. \SemiAUMCutoff{} remains a reasonable choice when substantial image corruption is expected and additional computation is available.

\addtocontents{toc}{\protect\setcounter{tocdepth}{-1}}
\clearpage
\section*{NeurIPS Paper Checklist}

\begin{enumerate}

\item {\bf Claims}
    \item[] Question: Do the main claims made in the abstract and introduction accurately reflect the paper's contributions and scope?
    \item[] Answer: \answerYes{} 
    \item[] Justification: Our claims made in the abstract and introduction match experimental results in the later sections.

\item {\bf Limitations}
    \item[] Question: Does the paper discuss the limitations of the work performed by the authors?
    \item[] Answer: \answerYes{} 
    \item[] Justification: Section~\ref{sec:conclusion} discusses the initial-label requirement, class-coverage sensitivity, additional SSL computation, and limits of generalization across pseudo-label generators; see also Appendices~\ref{appendix:pl_noise},~\ref{appendix:ssl_generality}, and~\ref{appendix:end_to_end_cost}.

\item {\bf Theory assumptions and proofs}
    \item[] Question: For each theoretical result, does the paper provide the full set of assumptions and a complete (and correct) proof?
    \item[] Answer: \answerNA{} 
    \item[] Justification: This paper does not include theoretical results.

    \item {\bf Experimental result reproducibility}
    \item[] Question: Does the paper fully disclose all the information needed to reproduce the main experimental results of the paper to the extent that it affects the main claims and/or conclusions of the paper (regardless of whether the code and data are provided or not)?
    \item[] Answer: \answerYes{} 
    \item[] Justification: The information needed to reproduce the experimental results is disclosed in Appendix~\ref{appendix:exp_details}.

\item {\bf Open access to data and code}
    \item[] Question: Does the paper provide open access to the data and code, with sufficient instructions to faithfully reproduce the main experimental results, as described in supplemental material?
    \item[] Answer: \answerYes{} 
    \item[] Justification: All experiments use publicly available datasets, models, and codebases. The setup directly follows these public implementations. Appendix~\ref{appendix:exp_details} specifies the choice of model, dataset, optimizers, and hyperparameters, which are sufficient to reproduce the experimental results. Our implementation and usage instructions are available at \url{https://github.com/cyseul/SemiPrune}; the experiments use the cited public datasets.
    

\item {\bf Experimental setting/details}
    \item[] Question: Does the paper specify all the training and test details (e.g., data splits, hyperparameters, how they were chosen, type of optimizer) necessary to understand the results?
    \item[] Answer: \answerYes{} 
    \item[] Justification: The experimental details are specified in Appendix~\ref{appendix:exp_details}.

\item {\bf Experiment statistical significance}
    \item[] Question: Does the paper report error bars suitably and correctly defined or other appropriate information about the statistical significance of the experiments?
    \item[] Answer: \answerNo{} 
    \item[] Justification: The main benchmark tables report mean downstream supervised training performance over five random seeds, using the same seeds across methods. Selected additional experiments report sample standard deviations, but the main benchmark tables omit error bars and we do not quantify variability across repeated SSL pseudo-label generation runs.

\item {\bf Experiments compute resources}
    \item[] Question: For each experiment, does the paper provide sufficient information on the computer resources (type of compute workers, memory, time of execution) needed to reproduce the experiments?
    \item[] Answer: \answerYes{} 
    \item[] Justification: We provide information on the computer resources in Appendix~\ref{appendix:computation_time}.
    
\item {\bf Code of ethics}
    \item[] Question: Does the research conducted in the paper conform, in every respect, with the NeurIPS Code of Ethics \url{https://neurips.cc/public/EthicsGuidelines}?
    \item[] Answer: \answerYes{} 
    \item[] Justification: The authors of this paper have reviewed the NeurIPS Code of Ethics. This paper conforms with the NeurIPS Code of Ethics in every respect.

\item {\bf Broader impacts}
    \item[] Question: Does the paper discuss both potential positive societal impacts and negative societal impacts of the work performed?
    \item[] Answer: \answerNA{} 
    \item[] Justification: This work is purely experimental. The main focus is on the suggestion of new methodology without a direct relation to downstream applications.
    
\item {\bf Safeguards}
    \item[] Question: Does the paper describe safeguards that have been put in place for responsible release of data or models that have a high risk for misuse (e.g., pre-trained language models, image generators, or scraped datasets)?
    \item[] Answer: \answerNA{} 
    \item[] Justification: This paper does not involve releasing any new data or models.

\item {\bf Licenses for existing assets}
    \item[] Question: Are the creators or original owners of assets (e.g., code, data, models), used in the paper, properly credited and are the license and terms of use explicitly mentioned and properly respected?
    \item[] Answer: \answerYes{} 
    \item[] Justification: We use publicly available datasets, optimizers, and pretrained models. All of them are properly cited in the paper.

\item {\bf New assets}
    \item[] Question: Are new assets introduced in the paper well documented and is the documentation provided alongside the assets?
    \item[] Answer: \answerNo{} 
    \item[] Justification: We release research code with environment setup and usage instructions at \url{https://github.com/cyseul/SemiPrune}. A dedicated asset card and complete licensing documentation are not provided.

\item {\bf Crowdsourcing and research with human subjects}
    \item[] Question: For crowdsourcing experiments and research with human subjects, does the paper include the full text of instructions given to participants and screenshots, if applicable, as well as details about compensation (if any)? 
    \item[] Answer: \answerNA{} 
    \item[] Justification: This paper does not involve crowdsourcing or research with human subjects.

\item {\bf Institutional review board (IRB) approvals or equivalent for research with human subjects}
    \item[] Question: Does the paper describe potential risks incurred by study participants, whether such risks were disclosed to the subjects, and whether Institutional Review Board (IRB) approvals (or an equivalent approval/review based on the requirements of your country or institution) were obtained?
    \item[] Answer: \answerNA{} 
    \item[] Justification: This paper does not involve crowdsourcing or research with human subjects.

\item {\bf Declaration of LLM usage}
    \item[] Question: Does the paper describe the usage of LLMs if it is an important, original, or non-standard component of the core methods in this research? Note that if the LLM is used only for writing, editing, or formatting purposes and does \emph{not} impact the core methodology, scientific rigor, or originality of the research, declaration is not required.
    \item[] Answer: \answerNA{} 
    \item[] Justification: LLM is used only for editing the paper.

\end{enumerate}
\end{document}